\documentclass{ieeetj}

\usepackage{enumerate}  
\usepackage{subcaption}
\usepackage{booktabs}
\usepackage[hypertexnames=false]{hyperref}
\usepackage{amsmath,mathrsfs}
\usepackage{amsthm}
\usepackage{amssymb}  
\usepackage{graphicx}
\usepackage{epstopdf}
\usepackage{epsfig}
\PassOptionsToPackage{hyphens}{url}
\usepackage{hyperref}
\usepackage{xcolor}
\usepackage{rotating}
\hypersetup{
    colorlinks,
    linkcolor={black},
    urlcolor={cyan!80!black}
}
\usepackage{bm}
\usepackage{algorithm}
\usepackage{algorithmic}
\usepackage{lipsum}
\usepackage{mathtools}
\usepackage{esvect}
\usepackage{afterpage}
\usepackage{scalerel}
\usepackage{hhline}
\usepackage{pdfpages}
\usepackage{glossaries}
\usepackage[acronym,style=super,nogroupskip,nonumberlist,toc=false]{glossaries-extra}
\usepackage{acro}
\DeclareAcronym{SITL}{
  short = SITL,
  long  = software-in-the-loop,
  short-indefinite = an,
  long-indefinite = a
}

\DeclareAcronym{FMU}{
  short = FMU,
  long  = flight management unit,
  short-indefinite = a,
  long-indefinite = a
}

\DeclareAcronym{CA}{
  short = CA,
  long  = control allocator,
  short-indefinite = a,
  long-indefinite = a
}

\DeclareAcronym{RTAI}{
  short = RTAI,
  long  = real-time application interface,
  short-indefinite = a,
  long-indefinite = a
}

\DeclareAcronym{dof}{
  short = DoF,
  long  = degrees-of-freedom,
  short-indefinite = a,
  long-indefinite = a
}

\DeclareAcronym{mocap}{
  short = MoCap,
  long  = motion capture system,
  short-indefinite = a,
  long-indefinite = a
}

\DeclareAcronym{NMPC}{
  short = NMPC,
  long  = Nonlinear Model Predictive Control,
  short-indefinite = an,
  long-indefinite = a
}

\DeclareAcronym{PWM}{
  short = PWM,
  long  = pulse width modulation,
  short-indefinite = a,
  long-indefinite = a
}

\DeclareAcronym{EKF}{
  short = EKF,
  long  = extended Kalman filter,
  short-indefinite = a,
  long-indefinite = a
}

\DeclareAcronym{RTOS}{
  short = RTOS,
  long  = real-time operating system,
  short-indefinite = a,
  long-indefinite = a
}

\DeclareAcronym{ISS}{
  short = ISS,
  long  = International Space Station,
  short-indefinite = a,
  long-indefinite = a
}

\DeclareAcronym{FSW}{
  short = FSW,
  long  = flight software,
  short-indefinite = a,
  long-indefinite = a
}

\usepackage[capitalize,nameinlink]{cleveref}
\usepackage{microtype}
\Crefname{equation}{Equation}{Equations}
\crefname{equation}{eq.}{eqs.}
\Crefname{figure}{Figure}{Figures}
\crefname{figure}{Fig.}{Figs.}
\crefname{table}{Tab.}{Tabs.}
\Crefname{table}{Table}{Tables}
\crefname{section}{Sec.}{Secs.}
\Crefname{section}{Section}{Sections}
\crefname{problem}{Problem}{Problems}
\Crefname{problem}{Problem}{Problems}
\crefname{definition}{Definition}{Definitions}
\Crefname{definition}{Definition}{Definitions}
\crefname{lemma}{Lemma}{Lemmas}
\Crefname{lemma}{Lemma}{Lemmas}
\crefname{theorem}{Thm.}{Thms.}
\Crefname{theorem}{Theorem}{Theorems}
\crefname{remark}{Rmk.}{Rmks.}
\Crefname{remark}{Remark}{Remarks}
\crefname{enumi}{item}{items}
\Crefname{enumi}{Item}{Items}
\crefname{algocf}{Alg.}{Algs.}
\Crefname{algocf}{Algorithm}{Algorithms}
\crefname{assumption}{Asm.}{Asms.}
\Crefname{assumption}{Assumption}{Assumptions}
\Crefname{Property}{Property}{Properties}
\crefname{Property}{Property}{Properties}

\AtBeginDocument{\RenewCommandCopy\qty\SI}
\usepackage{physics}
\usepackage[mode=text]{siunitx}
\DeclareSIUnit\bar{bar}
\usepackage{optidef}
\usepackage{tikz}
\usetikzlibrary{tikzmark,calc,decorations.pathreplacing}
\usetikzlibrary{shapes.geometric,positioning,quotes}
\tikzset{every edge quotes/.style =
          { fill = white,
            sloped,
            execute at begin node = $,
            execute at end node   = $  }}
\usepackage{textcomp}
\usepackage{anyfontsize}

\DeclareMathOperator*{\argmax}{arg\,max}

\newtheorem{definition}{\textbf{Definition}}

\newtheorem{remark}{\textbf{Remark}}

\newcommand{\myvar}[1]{{#1}}
\newcommand{\tildevar}[1]{\tilde{{#1}}}

\newcommand{\stepk}{{k}} 

\renewcommand{\vec}[1]{#1}
\newcommand{\vecs}[1]{{\vec{#1}}}  
\newcommand{\mat}[1]{#1}

\def\BibTeX{{\rm B\kern-.05em{\sc i\kern-.025em b}\kern-.08em
    T\kern-.1667em\lower.7ex\hbox{E}\kern-.125emX}}
\AtBeginDocument{\definecolor{tmlcncolor}{cmyk}{0.93,0.59,0.15,0.02}\definecolor{NavyBlue}{RGB}{0,86,125}}

\usepackage{placeins}

\newcommand{\FinalVersion}{}

\def\OJlogo{\vspace{-4pt}$<$Society logo(s) and publication title will appear here.$>$}
\def\seclogo{\vspace{10pt}$<$Society logo(s) and publication title will appear here.$>$}

\def\authorrefmark#1{\ensuremath{^{\textbf{#1}}}}

\begin{document}
\receiveddate{XX Month, XXXX}
\reviseddate{XX Month, XXXX}
\accepteddate{XX Month, XXXX}
\publisheddate{XX Month, XXXX}
\currentdate{XX Month, XXXX}
\doiinfo{XXXX.2022.1234567}

\markboth{}{Pedro Roque {et al.}}

\title{\LARGE 
Towards Open-Source and Modular Space Systems with ATMOS
}

\author{Pedro Roque\authorrefmark{1}, Sujet Phodapol\authorrefmark{1}, Elias Krantz\authorrefmark{2}, Jaeyoung Lim\authorrefmark{3}, Joris Verhagen\authorrefmark{4},\\ Frank J. Jiang\authorrefmark{1}, David Dörner\authorrefmark{2}, Huina Mao\authorrefmark{2}, Gunnar Tibert\authorrefmark{2}, Roland Siegwart\authorrefmark{3}, \\ Ivan Stenius\authorrefmark{2}, 
Jana Tumova\authorrefmark{4}, Christer Fuglesang\authorrefmark{2}, and Dimos V. Dimarogonas\authorrefmark{1}
}
\affil{Division of Decision and Control Systems, KTH Royal Institute of Technology, Stockholm, Sweden.}
\affil{School of Engineering Sciences, KTH Royal Institute of Technology, Stockholm, Sweden.}
\affil{Autonomous Systems Laboratory, ETH Z\"urich, Z\"urich, Switzerland. }
\affil{Division of Robotics, Perception and Learning, KTH Royal Institute of Technology, Stockholm, Sweden.}
\corresp{Corresponding author: Pedro Roque (email: padr@kth.se).}

\newcommand{\JFR}[0]{JFR}

\begin{abstract}
In the near future, autonomous space systems will compose many of the deployed spacecraft. Their tasks will involve autonomous rendezvous and proximity operations with large structures, such as inspections, assembly, and maintenance of orbiting space stations, as well as human-assistance tasks over shared workspaces. To promote replicable and reliable scientific results for autonomous control of spacecraft, we present the design of a space robotics laboratory based on open-source and modular software and hardware. The simulation software provides a software-in-the-loop architecture that seamlessly transfers simulated results to the hardware. Our results provide an insight into such a system, including comparisons of hardware and software results, as well as control and planning methodologies for controlling free-flying platforms.
\end{abstract}

\begin{IEEEkeywords}
Multi-Robot Systems, Orbital Robotics.
\end{IEEEkeywords}

\maketitle

\section*{Software and Hardware Releases}
Software and hardware contributions can be found in:
\begin{enumerate}
    \ifdefined\AnonVersion
    \item[1.] PX4Space and QGroundControl for PX4Space: 
    \par Included in the submission.
    \item[2.] Hardware release:
    \par \url{https://atmos.discower.io}
    \else
    \item[1.] PX4Space: 
    \par \url{https://github.com/DISCOWER/PX4-Space-Systems}
    \item[2.] QGroundControl for PX4Space:
    \par \url{https://github.com/DISCOWER/qgroundcontrol}
    \item[3.] ATMOS platform:
    \par \url{https://atmos.discower.io}
    \fi
\end{enumerate}

\section{Introduction}
\IEEEPARstart{T}{he} space sector has experienced significant growth in the last decade~\cite{villela2019towards}. Such growth is not only due to the decreased cost of access to space through multiple commercial operators~\cite{semanik2023}, but also due to the maturation of existing technologies and, consequently, reduced pricing for equipment. In the last twenty to thirty years, a few academic and industrial research facilities have been created to test space systems by replicating motion in microgravity on Earth. These facilities primarily rely on granite tables, resin floors, or other flat-calibrated surfaces such as optic tables.

\begin{figure}[tpb]
    \centering
    \includegraphics[width=\linewidth,clip,trim={0 3cm 0 2.5cm}]{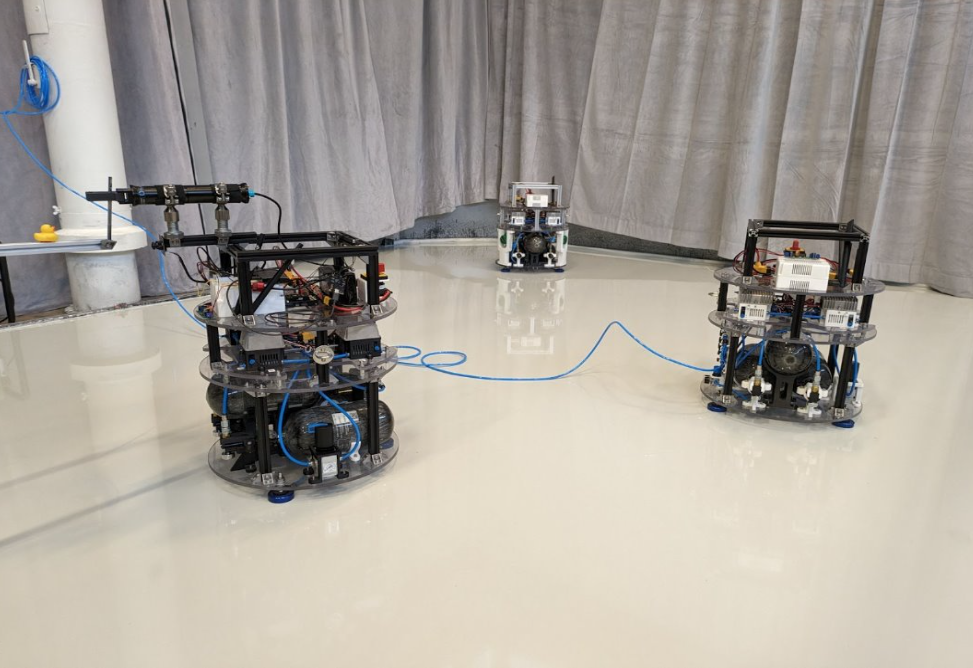}
    \caption{The KTH Space Robotics Laboratory, with three ATMOS free-flyers operating on a flat floor. One free-flyer is equipped with a manipulator payload, while another is connected to a low-pressure tether system.}
    \label{fig:facilities_overview}
\end{figure}

Some of the first microgravity testing facilities in Tohoku University~\cite{yoshida1994experimental,uyama2010integrated} and Tokyo University~\cite{murotsu1991experimental} used granite tables as calibrated flat surfaces where air-bearings supported robotic equipment mimicking undampened motion in the 2D plane. The National Technical University of Athens~\cite{papadopoulos2008ntua} proposed a similar test bed, where the platforms carry a microcontroller and an onboard computer that triggers $\text{CO}_2$-based thrusters. The facility also provides a vision-based ground truth that relies on fiducial markers for the position of each of the robotic platforms and sub-components. This facility was recently upgraded to more modern avionics, motion capture ground-truth positioning, and robotics communication software through the Robotics Operating System (ROS)~\cite{quigley2009ros}. Stanford University's Autonomous Systems Laboratory (previously maintained by the Aerospace Robotics Laboratory) free-flyer testbed~\cite{BonalliCauligiEtAl2019,StarekSchmerlingEtAl2016b,estrada2016free} uses a round platform as a free-flying robotic system for path planning, docking and capturing of space systems, paired with an open-source Python and ROS 2-based simulator. The open-source nature of the software allows other researchers to use the software to test their algorithms either in simulation only or as an intermediary step toward experiments in the facility. More recent granite table facilities in Europe are at the Space Research Center in Poland~\cite{basmadji2019microgravity} and at DLR Institute of Space Systems~\cite{daitx2016development}. On the latter, a 5 \ac{dof} platform provides attitude control in 3D and position control in 2D. The large granite surface comprises two granite slabs of \SI{4}{\metre}$\times$\SI{2.5}{\metre}, achieving a combined area of \SI{20}{\metre\squared}. The flatness of the surface is $20 \mu m$ edge-to-edge on each granite slab and $10\mu m$ table-to-table. The platform runs generated C code from Matlab~\cite{MATLAB} Simulink. The thruster systems are fed by \SI{300}{\bar} compressed air bottles, and an external ground-truth system is also available. The granite laboratory at NASA Ames, currently used to test the Astrobee free-flyer~\cite{bualat2015astrobee}, also relies on a granite surface to test the robotic vehicle that is currently operating on the \ac{ISS}. The free-flyer comprises two plenum cavities that provide electric propulsion to the spacecraft. Onboard, the platform has three ARM-based CPUs that control the platform actuation, provide state estimation from vision-based and inertial odometry, and three payload bays for guest science research. The software stack is open-source with a complete simulation package running in ROS and Gazebo~\cite{Koenig-2004-394}, making it easy for anyone to test their algorithms onboard the Astrobee.

Despite the considerable precision that can be achieved with granite tables, we identify two drawbacks of this approach: cost and weight. Typically, granite tables are limited to ground-floor installations due to the pressure they apply on the supporting surface. Moreover, large granite slabs are also costly, and aggregating multiple units is also expensive in maintaining the operation, as they need to be calibrated in periodic intervals. An alternative to this solution is to use resin floors. The Space Robotics Laboratory at the Naval Postgraduate School~\cite{romano2007laboratory} uses a flat floor of approximately \SI{20}{\metre\squared} built with epoxy resin and initially developed to test docking scenarios for two spacecraft that were purpose-built for this facility. This system provides a Pentium III in a PC/104 compatible avionics stack and uses cold-gas thrusters as actuators. Here, localization is done via an indoor GPS system, from which the vehicle position is obtained through triangulation to two fixed transmitters. In~\cite{barrett2009demonstration}, a facility at the University of South California showcases a resin floor capable of operating multiple 3 \ac{dof} platforms with vision-based docking mechanisms built with commercial off-the-shelf components. Similar facilities have been proposed by the University of Kentucky Aerospace Deployment Dynamics Laboratory and ATK Robotic Rendezvous and Proximity (RPO) testing facility~\cite{atk_rpo}. With the same operating principles, the facilities at Georgia Institute of Technology~\cite{tsiotras2014astros} and Rensselaer Polytechnic Institute~\cite{saulnier2014six} use resin floors and air bearings for a frictionless motion of their platforms, which can provide 5 and 6 \ac{dof}, respectively. Although the 5 \ac{dof} system in~\cite{tsiotras2014astros} is similar to the one in~\cite{daitx2016development}, the 6 \ac{dof} system in~\cite{saulnier2014six} augments the 5 \ac{dof} system with an elevator that can vertically move an attitude-controlled stage. This system can mimic the 6 \ac{dof} motion at the expense of some imperfection on the Z axis at high speeds. Both platforms use PC/104 computers, with~\cite{tsiotras2014astros} using Matlab and Simulink interfaces, while \cite{saulnier2014six} uses a \ac{RTAI} Linux operating system. The largest resin-based floor system is in the California Institute of Technology~\cite{nakka2018six}. This facility provides a more than \SI{40}{\metre\squared} area with multiple platforms that can provide 3 to 6 \ac{dof}. The propulsion method uses compressed air tanks at a \SI{300}{\bar}, and the onboard computer is an Nvidia Jetson TX2, which contains an integrated GPU for onboard parallel computations. In Europe, multiple facilities have also been created. The European Space Agency Orbital Robotics laboratory~\cite{kolvenbach2016recent} provides a similar facility composed mainly of 3 \ac{dof} platforms using compressed air as a propellant. Two other recent facilities are the ones in the Luleå University of Technology~\cite{banerjee2021slider,nieto2021concurrent} and in the University of Luxembourg~\cite{olivares2023zero}. These facilities provide access to 3 Degree-of-Freedom platforms and motion capture systems for ground truth. 

Other approaches to recreate microgravity motion have used optic~\cite{miller2000spheres,sabatini2023facility,rughani2019swarm} and pressurized~\cite{trentlage2018elissa} tables. 
In particular, the facility at the Massachusetts Institute of Technology Space Systems Laboratory~\cite{miller2000spheres} has been used both to create and to serve as a ground testing facility for the MIT Synchronized Position Hold Engage and Reorient Experimental Satellite (SPHERES), a set of three satellite demonstrators that operated inside the ISS from May 2006 to December 2019. The SPHERES used $\text{CO}_2$ as their main propellant. The onboard computer consisted of a Texas Instruments DSP programmed in C. Additionally, a simulator in Matlab and Simulink was available. The optic table provided a \SI{2.7}{\metre\squared} operating surface, which needed to be calibrated regularly to ensure accurate results. The facilities in \cite{sabatini2023facility,rughani2019swarm}  use similar optic tables. These tables are usually paired with a glass panel to provide a smooth operating surface for air bearings; therefore, their precision depends on the glass panel's properties. The ELISSA facility described in \cite{trentlage2018elissa} provides an operation method that works mostly in an opposite fashion to most of the systems described here. In this case, the air cushion is created by small nozzles distributed on the table surface and a piece of acrylic or glass that is attached to the bottom part of the simulated spacecraft. This allows the operation time of simulated spacecraft to not depend on the available air supply but only on its battery. The NASA Johnson Space Center Precision Air Bearing Floor \cite{NASA_PABF} is capable of simulating human-scale platforms for training astronauts for extravehicular activities in microgravity. The NASA Jet Propulsion Lab GSAT laboratory \cite{sternberg2018jpl,rivera2023multi} supports multiple spacecraft analogue systems on their flat floor, composed by individual, calibrated plates of aluminium. Lastly, the LASR \cite{adams2022velocimeter} laboratory at Texas A\&M supports a large flat floor facility with a total of 2000 square feet and simulates satellite dynamics using holonomic wheeled robots instead of air bearings.
Multiple other microgravity simulator facilities can be found in~\cite{rybus2016planar,papadopoulos2021robotic,WILDE2019100552,flores2014review,menon2007issues,osuman2010multi,pronk1996flat,schwartz2003historical}. 

Despite numerous platforms and laboratory facilities, we found two significant problems related to platform replication, applicability of research results, and benchmarking. First, most software is not open-source or relies on custom-developed low-level hardware, making it hard to replicate the testing conditions. Secondly, platform modularity is often not considered; thus, making adjustments or future-proofing challenging. Due to these reasons, transitioning from ground testing to orbit testing is difficult outside of the proposed facilities. This article proposes an open-source microgravity simulation laboratory to tackle these drawbacks. For this facility, we propose and develop a 3 \ac{dof} Autonomy Testbed for Multi-purpose Orbital Systems (ATMOS) platform with open hardware and software to facilitate their replication at a low cost. The hardware is based on commercial off-the-shelf components that are widely available. The low-level microcontroller is based on the Pixhawk 6X Mini, while the high-level computer is an Nvidia Jetson Orin NX. This combination allows us to: i) run \emph{PX4Space}, a branch of the open-source PX4 we developed for thruster-based systems with \ac{SITL} capabilities, ii) support multiple payloads with heterogeneous power and communication needs, and iii) be compatible with on-orbit facilities software stacks, such as the Astrobee \ac{FSW}\cite{fluckiger2018astrobee}, aiming at reducing the time-to-orbit of ground experiments.
The free-flying platform operates on top of three air bearings with a maximum combined payload of approximately \SI{150}{\kg}. The air bearings and thrusters are air operated from three \SI{1.5}{\liter} bottles, filled at \SI{300}{\bar}, and separately regulated to \SI{6}{\bar} for the actuation and air bearing systems. The platform contains a total of 8 thrusters actuated via \ac{PWM}. The facility provides access to low- and high-pressure compressors. The low-pressure compressor can continuously operate three tethered platforms with an uninterrupted air supply. At the same time, the high-pressure compressor is mainly used for fast experimental turnover time by rapidly refilling the compressed air bottles. The operational area has an approximate dimension of \SI{15}{\metre\squared} covered by a Qualisys active motion capture system that provides an accurate, sub-millimeter ground-truth position of all tracked rigid bodies. 

The remainder of the article is divided as follows: in \cref{sec:facilities}, we introduce the available facilities. \Cref{sec:hardware,sec:software} detail the hardware and software of the proposed modular and open-source platform, ATMOS, while \cref{sec:autonomy} discusses its autonomy capabilities. Ending the manuscript, \cref{sec:prel_results} provides preliminary results of the proposed hardware and software package along with a small discussion on these. \Cref{sec:discussion} concludes the article.

\textit{Notation:} Matrices are denoted by capital letters. Let
${a}^T$ be the transpose of ${a}$. The orthogonal basis vectors of a frame $\mathcal{A}$ are denoted $\{\vec{a}_x,\vec{a}_y,\vec{a}_z\}$.
The inertial reference frame is generally omitted. Sets are defined in blackboard bold, $\mathbb A$. The weighted vector norm $\sqrt{\vec x^T \mat A \vec x}$ is denoted $\norm{\vec x}_{\mat A}$.

\section{Facilities}
\label{sec:facilities}
This chapter provides a detailed overview of the laboratory facilities. These encompass the resin flat floor, the motion capture system, and the compressors for high and low pressures. \Cref{fig:facilities_overview} shows the free-flying platforms available at the facility.

\begin{figure}[tpb]
    \centering
    \subfloat[Measurements performed after the initial resin pour.
    ]
    {
    \includegraphics[width=.95\linewidth]{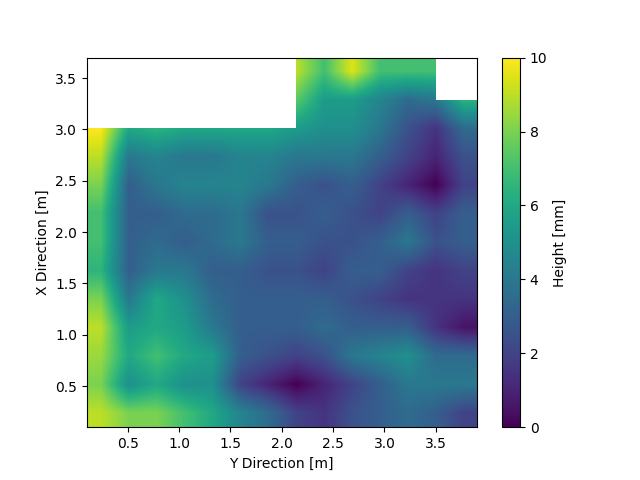}
    \label{exp:floor_meas_1}
    } \\
    \subfloat[Measurements performed after sanding and reapplying resin.
    ]
    {
    \includegraphics[width=.95\linewidth]{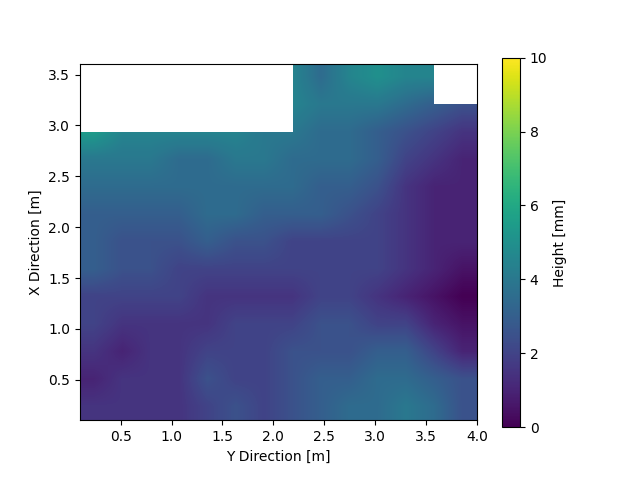}
    \label{exp:floor_meas_2}
    } 
    \caption{Floor level, in millimeters, after the first and second resin pour. Between pourings, the floor was measured and leveled through sanding. The measurements were obtained in a \SI{30}{\cm} grid, and the plots show a bilinear interpolation from these measurements.}
    \label{fig:floor_levels}
\end{figure}

\subsection{Epoxy Floor}

The flat floor depicted in \cref{fig:facilities_overview} was developed over one year with multiple resin layers. In between each pour, measurements were collected with a leveling laser, and the higher areas were sanded for more extended periods than the lower areas. In \cref{fig:floor_levels}, we show the floor levelness after the first and second pours of epoxy resin.
Currently, the flat floor provides a maximum declination of \SI{2}{\milli\metre\per\text{m}}. Although a higher precision can be achieved with granite surfaces, the resin floor provided a more cost-efficient installation. We must note that more accurate solutions can be obtained with different types of resins, as also seen in \cite{nakka2018six}, and that our platform has sufficient thrust to compensate for such disturbances.

\begin{figure}[tpb]
    \centering
    \subfloat[Three out of six Qualisys M5 motion capture cameras overlooking the experimental area.]{
    \includegraphics[width=.95\linewidth]{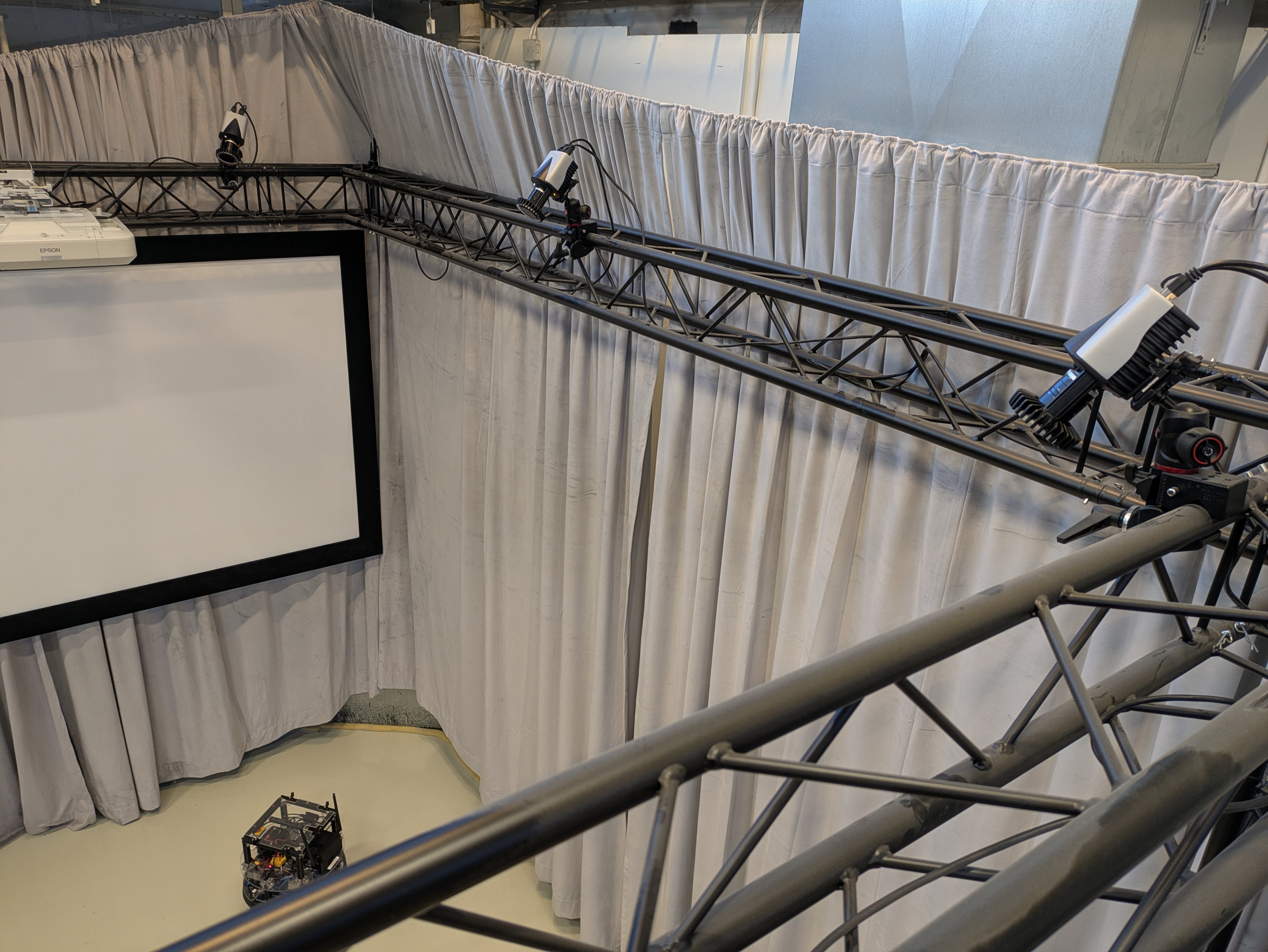}
    \label{exp:mocap_cameras}
    } \\
    \vspace{2mm}
    \subfloat[Qualisys' Naked Traqr active marker system attached to the platform.]{
    \includegraphics[width=.95\linewidth]{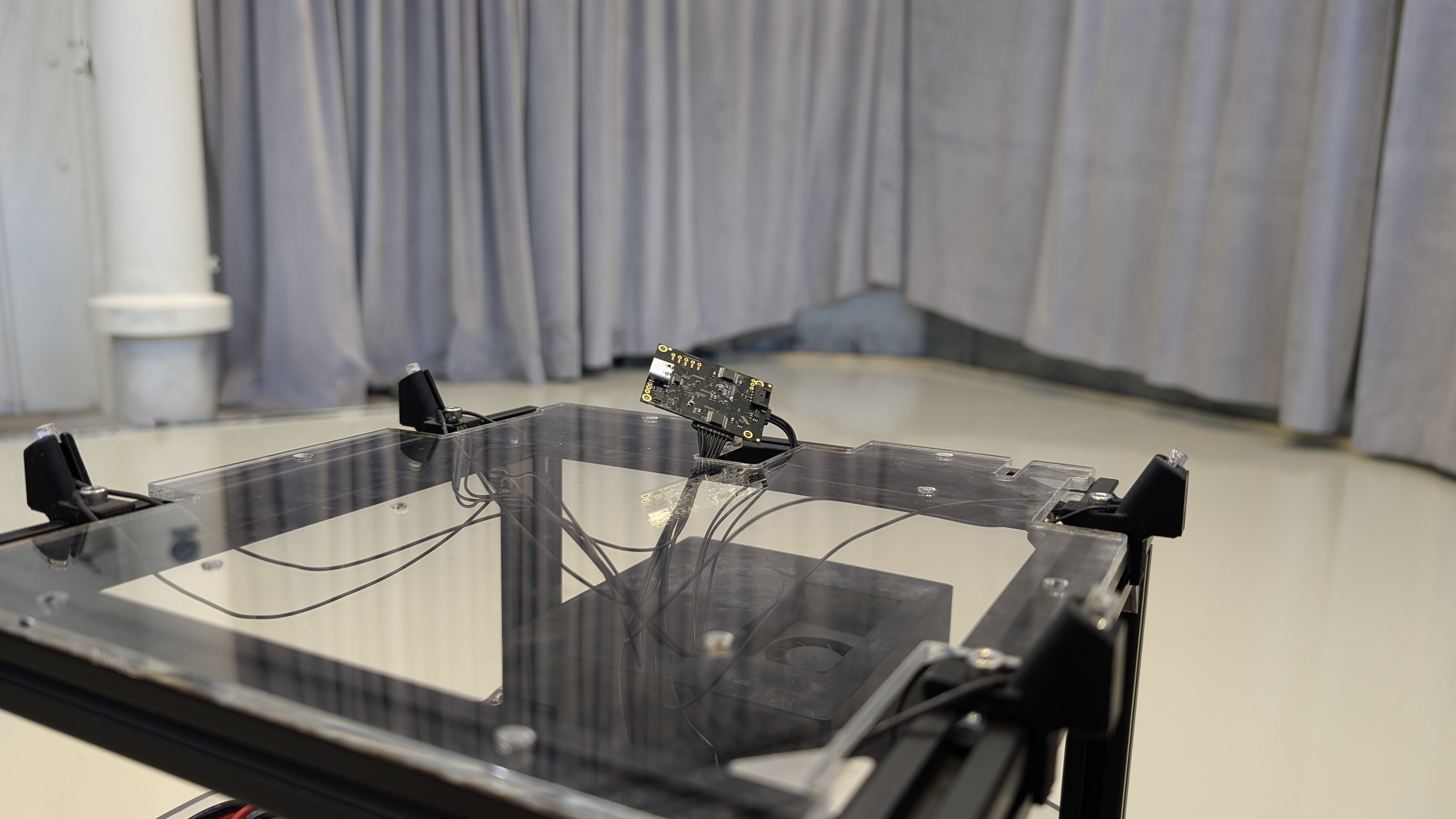}
    \label{exp:mocap_traqr}
    } 
    \caption{The motion capture system is composed of six Qualisys M5 cameras - three are shown in (a) - and an active LED system onboard the free-flying platform, seen in (b). Each LED flashes at a unique frequency captured passively by the cameras, identifying both the LED and the rigid body pose in real-time. }
    \label{fig:mocap_and_tracker}
\end{figure}

\subsection{Motion Capture System}

A \ac{mocap} provided by Qualisys with six cameras is installed above the operating area shown in \cref{fig:facilities_overview}. These cameras offer sub-millimeter tracking of any rigid body in the workspace at a frequency of \SI{100}{\hertz}. Since the floor is highly reflective due to the polishing of the surface, the motion capture system operates in active mode, where the cameras passively capture the infrared LEDs on the platform. The infrared LEDs strobe at distinct frequencies and, therefore, are identifiable by the system. The LEDs are connected to a Qualisys Naked Traqr that controls the frequency of each LED. In~\cref{fig:mocap_and_tracker}, we show both the motion capture system cameras and the active tracking unit onboard one of our free-flying platforms.

\begin{figure}[tpb]
    \centering
    \includegraphics[width=0.95\linewidth]{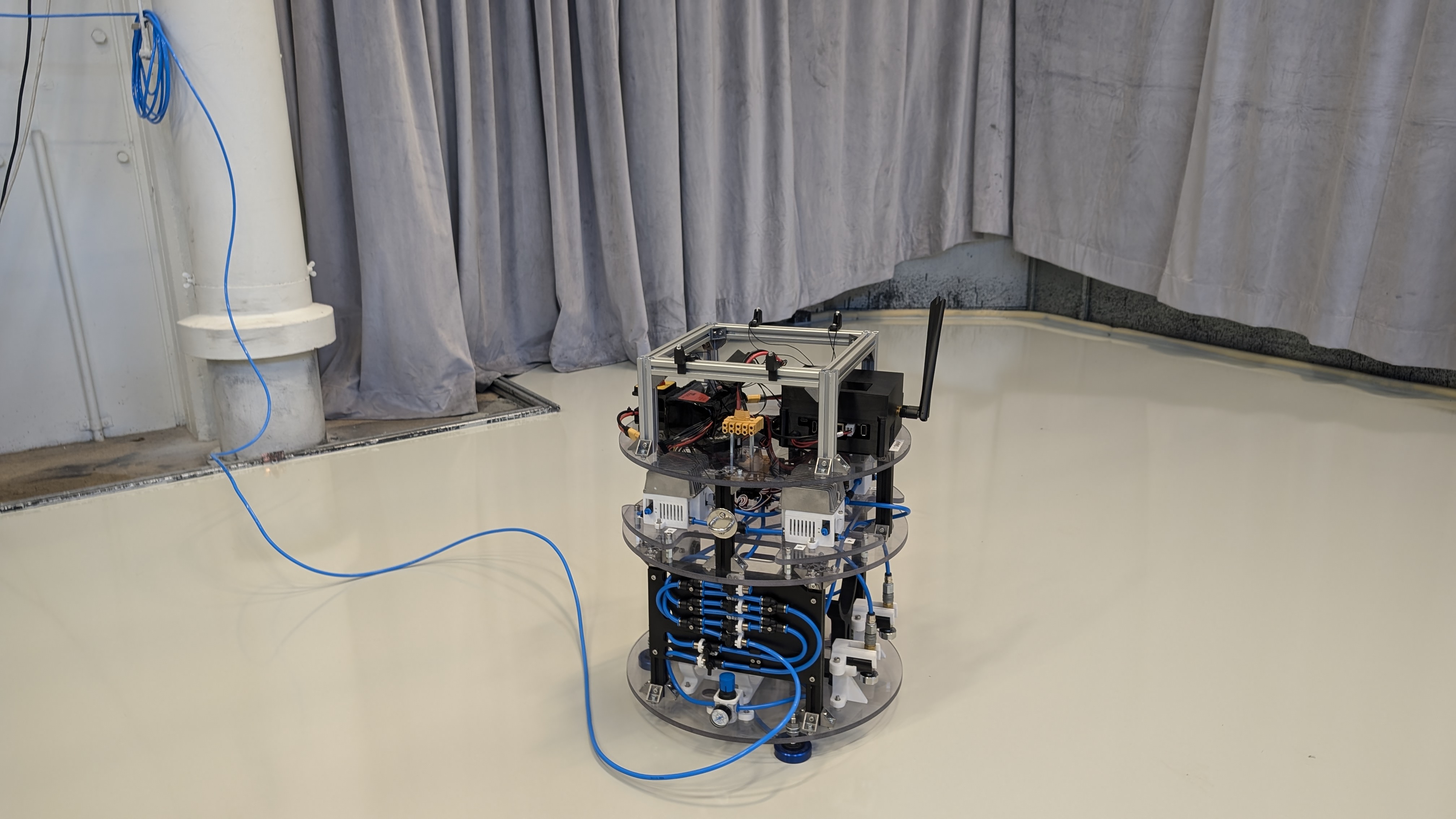}
    \caption{Low pressure (\SI{10}{\bar}) tether cable attached to one of the free-flying platforms. The tether cable can provide continuous, uninterrupted operation for unlimited time.}
    \label{fig:low_pressure_compressor}
\end{figure}

\subsection{Pressurized Air Supplies}

Low and high-pressure compressors are two critical pieces of equipment that can sustain prolonged test sessions at the laboratory. Our facility has access to an Atlas Copco \SI{10}{\bar} compressor capable of simultaneously providing tethered air supply to three of our platforms. The air supply is given through tether cables attached to the pressurized manifold on the platform. Then, the pressurized air is regulated to \SI{6}{\bar} to feed the air bearings and the thruster actuation system. An image of the tether attached to a robot is seen \cref{fig:low_pressure_compressor}. 
The high-pressure compressor unit consists of a Bauer PE250-MVE \SI{300}{\bar} compressor paired with a \SI{50}{\liter} reservoir. It also includes a remote filling station for quickly refilling pressurized air bottles used during tests. 

As the tether cable induces external disturbances on the platform, it is primarily used during development and testing sessions. When untethered operation is required, the three \SI{1.5}{\liter} bottles onboard the platform are used instead. These bottles can be rapidly refueled before each experiment using the high-pressure compressor.

\section{Free-flyers Hardware}
\label{sec:hardware}
A core part of our space robotics laboratory are the ATMOS free-flyers, which simulate the dynamics of a small satellite operating in microgravity. To our knowledge, such platforms are unavailable for off-the-shelf purchase due to each laboratory's different needs. In this section, we detail the hardware needed to build our modular free-flying platforms, proposing a flexible platform that other space robotics facilities can use.

\begin{figure}[tpb]
    \centering
    \includegraphics[width=0.95\linewidth]{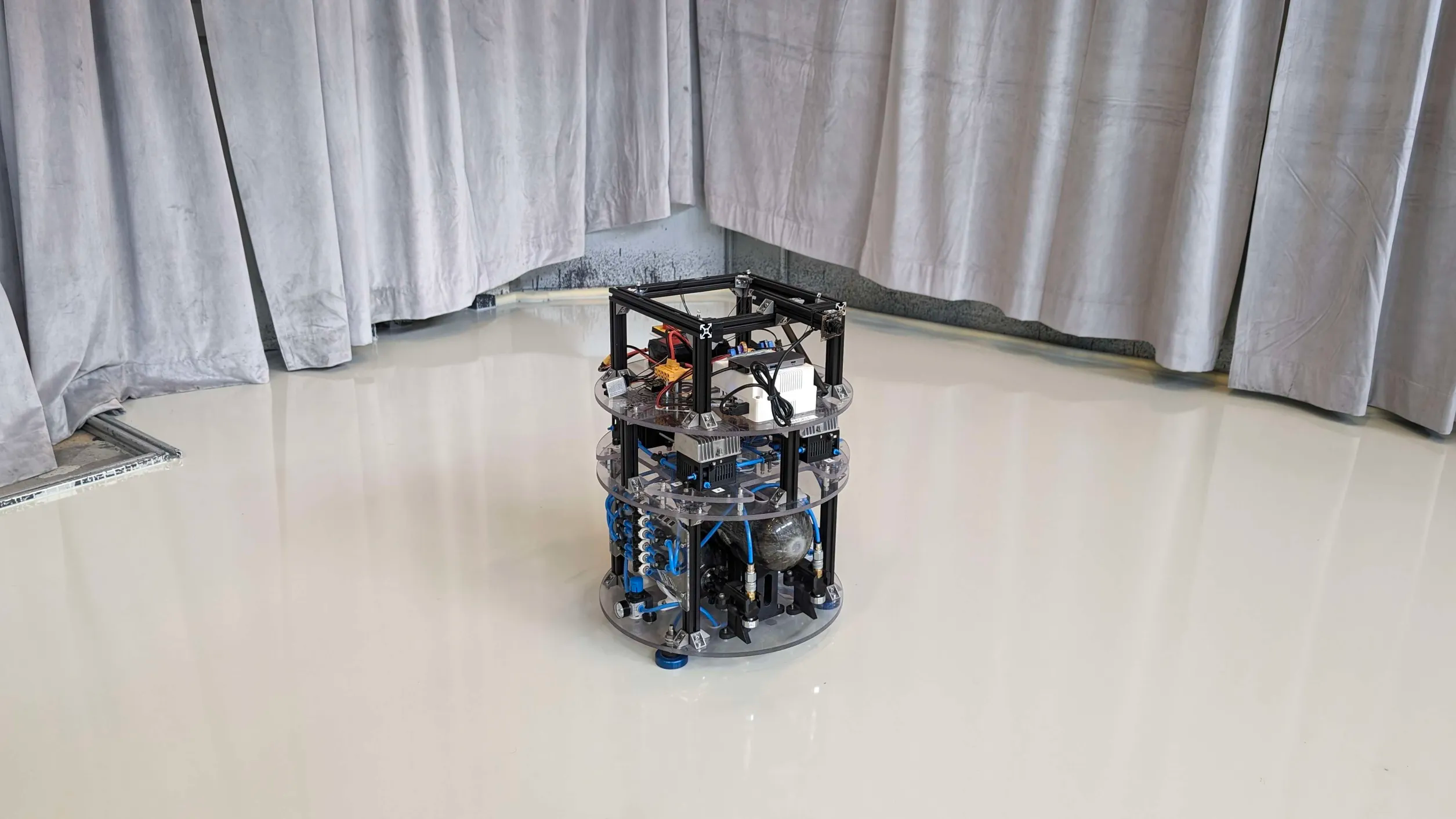}
    \caption{The ATMOS free-flyer. The platform is \SI{40}{\cm} wide and approximately \SI{50}{\cm} tall, with four sections: a pressurized section, an actuation section, an electronics section, and a payload section.}
    \label{fig:freeflyer_picture}
\end{figure}
\begin{figure}[tpb]
    \centering
    \includegraphics[width=0.95\linewidth,clip,trim={2cm 5cm 2cm 3cm}]{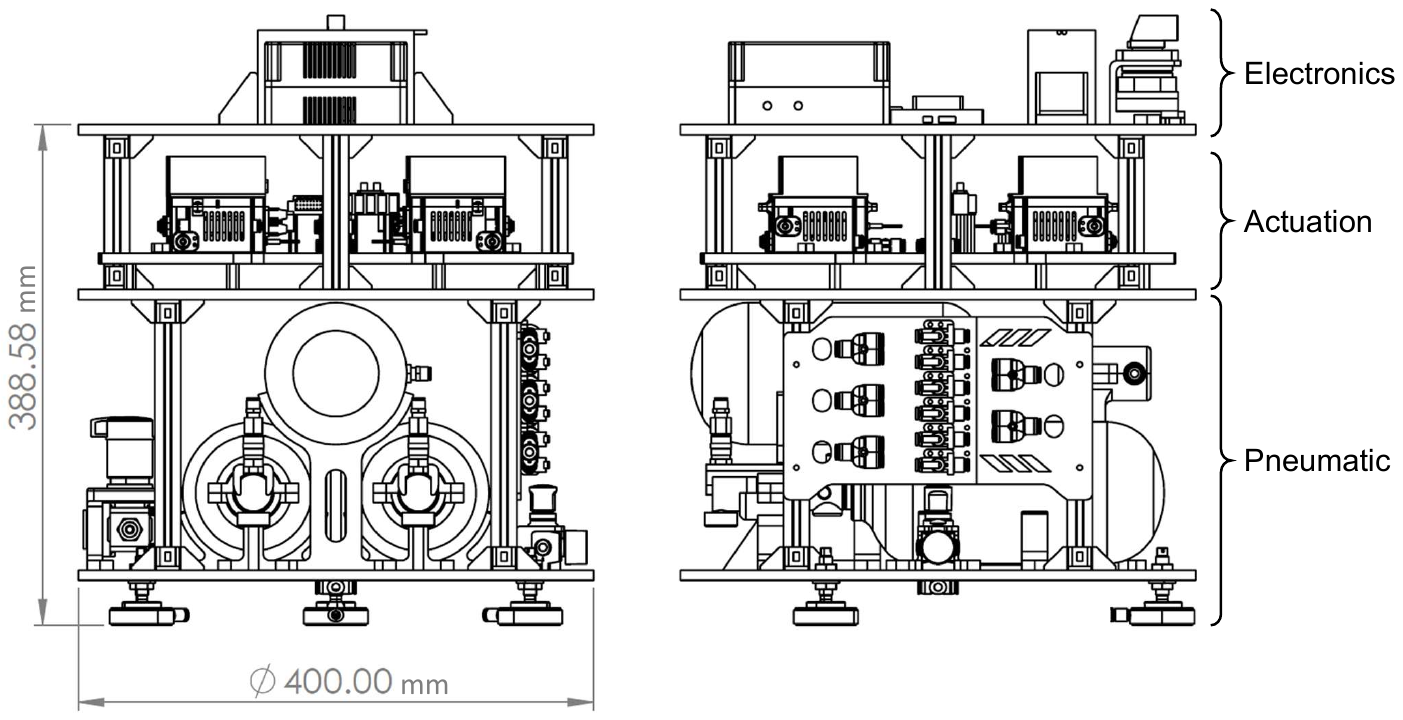}
    \caption{Schematic overview of the platform. The dimensions of the free-flyer are shown on the left, while three of the four layers that compose the robotic agent are shown on the right.}
    \label{fig:schematic_overview}
\end{figure}

\subsection{Hardware Overview}

The free-flyers developed for the laboratory have the following design goals: i) modularity, allowing for easy module substitution and template designing without significant changes to the platform; ii) cost-efficiency, allowing for an economical replication of these units, and iii) guest science support, providing a facility that external researchers or industry partners can use as a payload bearer to test hardware or software in the laboratory. With these goals in mind, the final design of our platform is shown in \cref{fig:freeflyer_picture}.

The dimensions of the platform, as well as a schematic overview, are available in \cref{fig:schematic_overview}.
An overview of the platform actuation and inertial parameters is shown in \cref{tab:inertial_parameters}. We compare our platform with the NASA Astrobee robot, a widely used platform for guest science research aboard the \ac{ISS}. The information regarding the Astrobee platform was collected from \cite{albee2022online,fluckiger2018astrobee}.

\begin{table}[h]
    \caption{Free-flyer inertial and actuation parameters based on measured weight and estimated inertial matrix from computer-aided design (CAD). We compare our platform with the Astrobee facility in both ground (G) and flight versions (F).}
    \label{tab:inertial_parameters}
    \begin{tabular}{llll}
    \toprule
    Parameters       & ATMOS & Astrobee F  & Astrobee G  \\
    \midrule
    Mass [\SI{}{\kilogram}]                             & 16.8 & 9.58    & 18.97   \\
    Moment of inertia [\SI{}{\kilogram\metre\squared}]           & 0.297 & 0.162   & 0.252  \\
    Max. thrust (x-axis) [\SI{}{\newton}]              & 3.0 & 0.849   & 0.849    \\
    Max. thrust (y-axis) [\SI{}{\newton}]              & 3.0 & 0.406   & 0.406    \\
    Max. torque [\SI{}{\newton\metre}]                      & 0.51 & 0.126   & 0.126   \\ 
    \bottomrule
    \end{tabular} 
\end{table}

\subsection{Pneumatic Section}
\begin{figure}[t]
    \centering
    \includegraphics[width=\linewidth,clip,trim={1.5cm 3cm 5.8cm 2cm}]{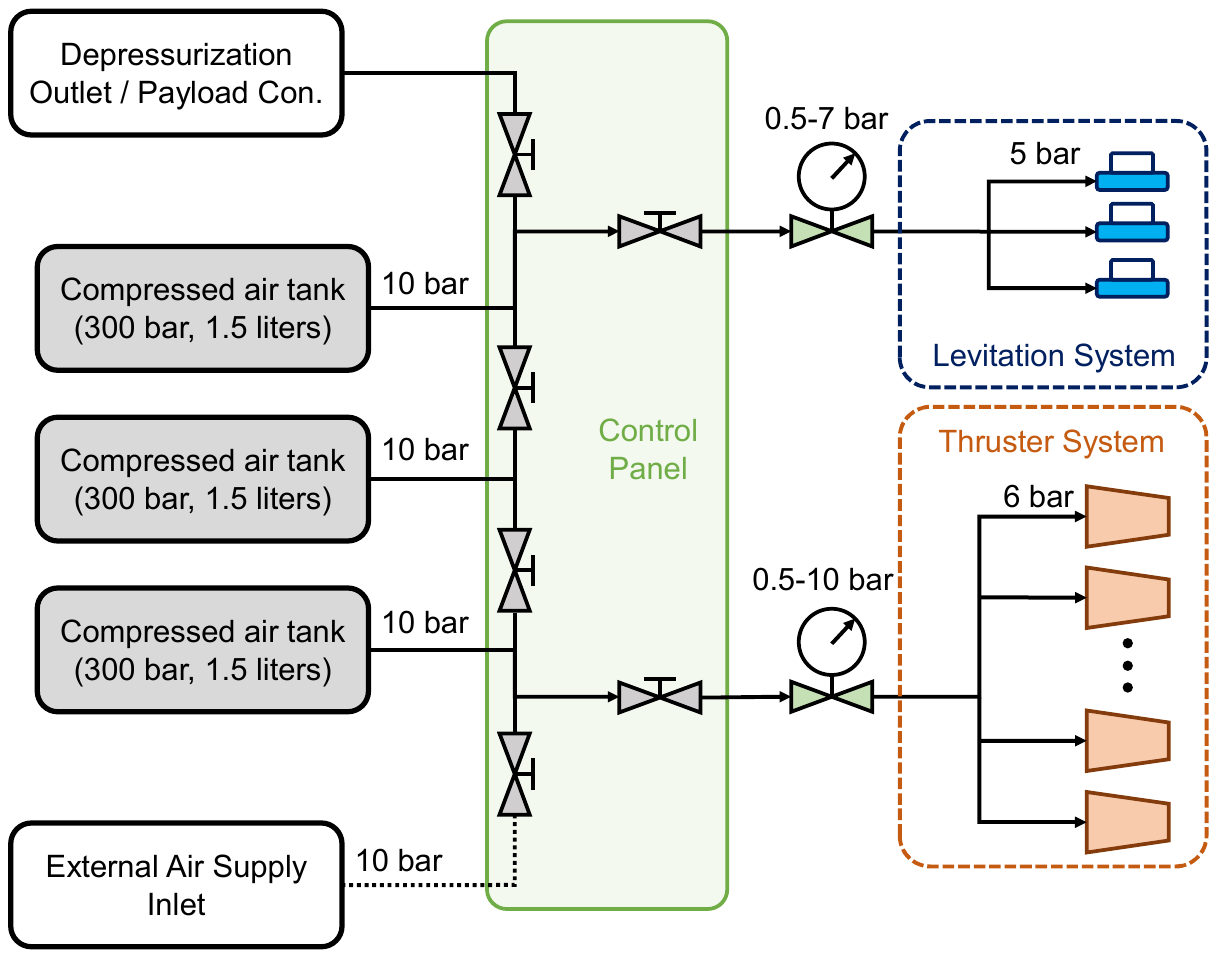}
    \caption{Pressurized section diagram. The air supply is provided by \SI{1.5}{\liter} compressed breathable air bottles with a filled pressure of \SI{300}{\bar}. Attached to the bottles, two sequential pressure regulators output \SI{55}{\bar} to the pressurized manifold. The manifold can select the number of bottles providing air to both the actuation and floating subsystems. Each subsystem has a pressure regulator capable of providing $0.5$ to \SI{10}{\bar} of pressurized air. Lastly, the manifold also provides an optional external pressurized air supply inlet for tethered operation and a depressurization/payload connection outlet.}
    \label{fig:pressurized_section}
\end{figure}

The pressurized section provides all sub-components with the necessary capabilities for frictionless motion, propulsion, and pressure regulation. A representation of this section is shown in \cref{fig:pressurized_section}.

\begin{figure}[t]
   \centering \includegraphics[width=0.95\linewidth,clip,trim={0.5cm 0cm 0cm 3cm}]{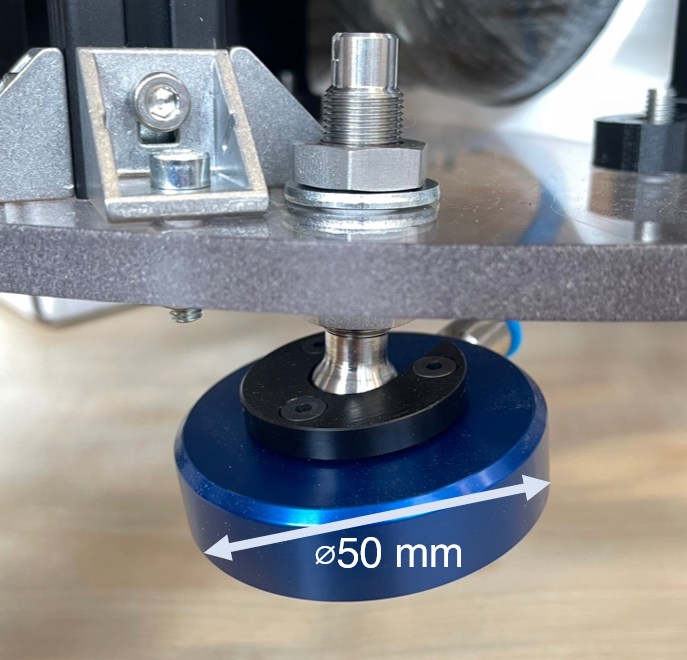}
   \caption{NewWay S105001 air bearing at the base of ATMOS.}
   \label{fig:bearing}
\end{figure}

Three bottles provide breathable compressed air at a pressure of \SI{300}{\bar} with a combined volume of \SI{4.5}{\liter}. Two bottle-attached regulators then regulate the high pressure down to \SI{10}{\bar}. Their description is given in \cref{tab:pneumatic_components}.

The three \SI{50}{\mm} air bearings at the bottom of the platform create an air cushion of $15$ to \SI{20}{\micro\metre}, thus enabling the frictionless motion of the platform without any contact to the floor. \Cref{fig:bearing} pictures the air bearings, which require an input pressure of \SI{5}{\bar} and can hold a load up to \SI{52}{\kg} each. Combining the three bearings on a single platform, a maximum load of approximately \SI{156}{\kg} is achieved.

\begin{table}[h]
    \begin{center}
        \caption{Pneumatic components for the ATMOS free-flyer.}
        \label{tab:pneumatic_components}
        \begin{tabular}{ll}
        \toprule
        \multicolumn{1}{l}{Component}        & \multicolumn{1}{l}{Model}                              \\
        \midrule
        Compressed air tank (\SI{300}{\bar})            & DYE CORE AIR TANK 1.5L 4500PSI              \\
        Bottle regulator (to \SI{55}{\bar})               & DYE LT Throttle Regulator 4500PSI \\
        Bottle regulator (to \SI{10}{\bar})              & Polarstar micro MR GEN2 Regulator           \\
        Floating regulator                     & MS2-LR-QS6-D6-AR-BAR-B   \\
        Thrusters regulator                     & MS4-LR-1/4-D7-AS                 \\ 
        \bottomrule
        \end{tabular}
    \end{center}
\end{table}

\subsection{Actuation Modules}

For our platform, two actuation modules were designed: i) a solenoid valve actuation plate, referred to as a thruster plate, with a total of eight thrusters that mimic the Reaction Control System (RCS) on a spacecraft; and ii) a propeller-based actuation plate that aims at mimicking the actuation dynamics of the free-flyers such as the NASA Astrobee\cite{bualat2015astrobee}, with a total of four motors capable of bi-directional rotation. These two actuation plates also allow us to have two actuation modalities: the thruster plate only allows for maximum thrust / no thrust with \ac{PWM}, while the propeller plate allows for selecting any target thrust within the control set. 

An essential aspect of the platform is the modularity of the actuation modules, making it possible for other system users or laboratories to share their designs and augment the capabilities of ATMOS.

\begin{figure}[t]
   \centering
   \includegraphics[width=1\linewidth,clip,trim={5cm 0 5cm 0}]{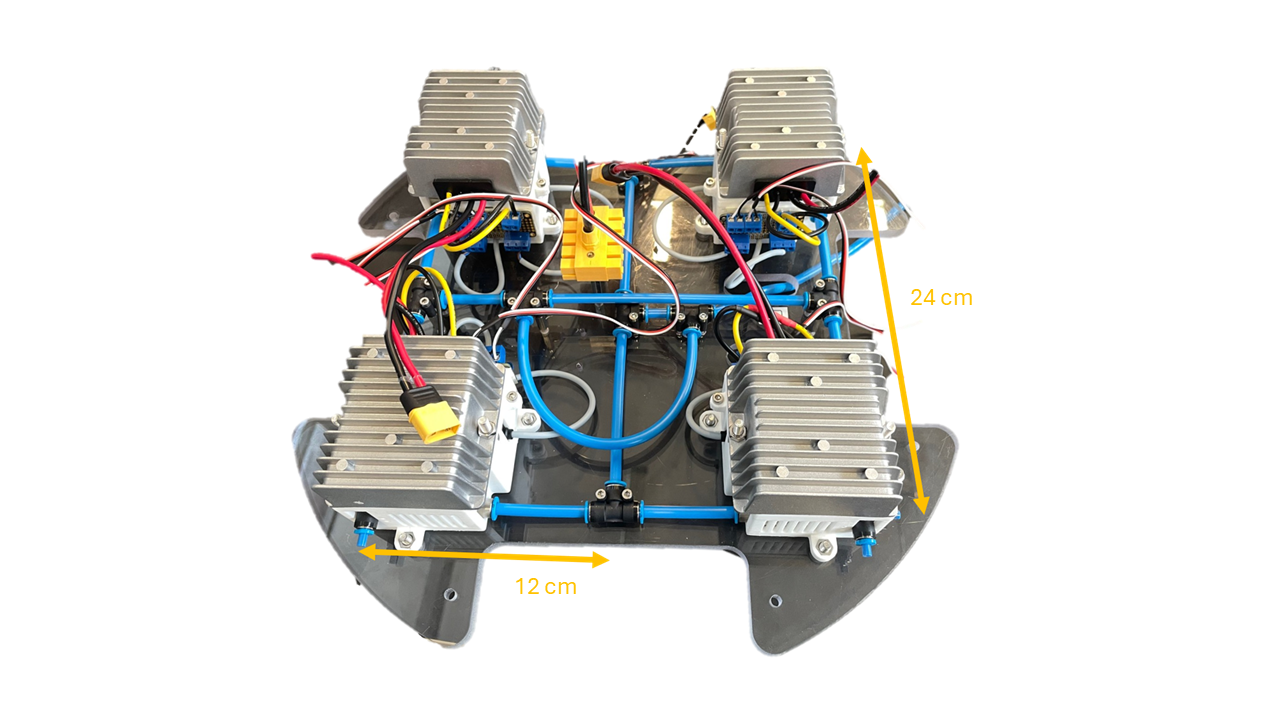}
   \caption{The thruster plate comprises eight thrusters divided into four modules. The modules are shown in \cref{fig:thruster_module}. The thruster plate interfaces with the electronics via one cable for power and one 8-in-1 cable for PWM signaling.}
   \label{fig:thruster_configuration}
\end{figure}

\begin{figure}[t]
   \centering
   \includegraphics[width=\linewidth,clip,trim={1cm 4cm 1cm 3cm}]{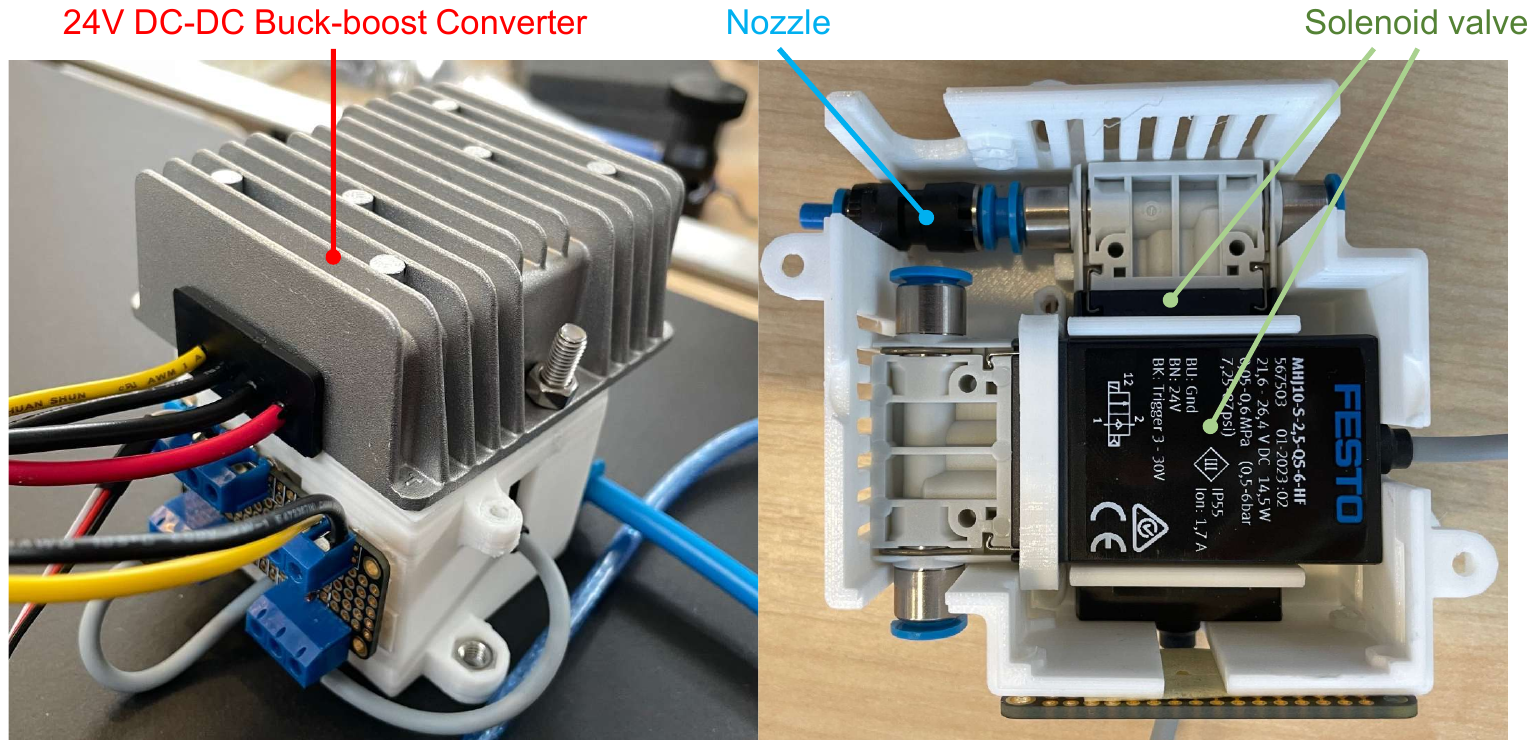}
   \caption{Each thruster module consists of two solenoid valves with integrated electronics, two nozzles, and one buck-boost converter and interfaces with the platform via two signal cables and one battery power connection. In the image to the right, we can see the orthogonal placement of the solenoid valves and one nozzle.}
   \label{fig:thruster_module}
\end{figure}

The thruster plate has eight solenoid valves divided into four modules, each with a pair of thrusters. The modules are placed in the vertices of a square with \SI{24}{\cm} edges as shown in \cref{fig:thruster_configuration}. This configuration allows the platform to be holonomic in the motion plane, with 3 \ac{dof}, while providing thruster redundancy for up to two non-collinear thruster failures. The solenoid valves provide a maximum switching frequency of \SI{500}{\hertz}. The platforms are nominally configured in software to a fixed switching frequency of \SI{10}{\hertz} with a controllable duty cycle from $0$ to \SI{100}{\ms}. For compactness, we include two solenoids and one \SI{24}{\V} buck-boost converter into one thruster module, shown in \cref{fig:thruster_module}. Compressed air is supplied via the onboard manifold described in \cref{fig:pressurized_section} and the blue tubing shown in \cref{fig:thruster_configuration}.


\begin{figure}[tpb]
   \centering
   \includegraphics[width=1\linewidth]{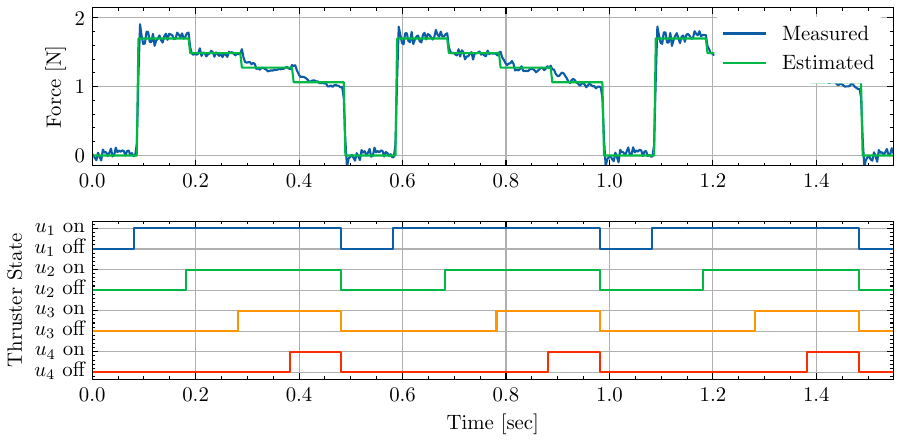}
   \caption{Force measurement of one thruster when operating with up to three additional thrusters, each one reducing the achieved force by approximately \SI{12.5}{\percent}.}
   \label{fig:4valves_thrust_measurement}
\end{figure}

The thrusters operate at \SI{6}{\bar}, drawing air through \SI{4}{\mm} diameter hoses and expelling it via \SI{2}{\mm} nozzles. Experimental measurements show that a single thruster in this system generates \SI{1.7}{\newton} of force while consuming \SI{3.5}{\gram\per\second} when open; closely matching the theoretical maximum of \SI{1.76}{\newton} for isentropic nozzle flow under these conditions. However, since the thrusters share a common air supply, interactions between them are inevitable. This coupling effect was measured and is illustrated in \cref{fig:4valves_thrust_measurement}, where the force output of one thruster is recorded while up to three additional thrusters are simultaneously active. During this measurement, all thrusters ran at the nominal \SI{10}{\hertz} frequency whilst switching between $0$ and \SI{100}{\percent} duty cycle. This data was collected at $320$ samples/second using a TAL220 10kg load cell amplified with a NAU7802 analog-to-digital converter. Each additional thruster reduces the output force of the measured thruster by approximately \SI{12.5}{\percent}. It should be noted that this is a worst-case scenario, where several thrusters are operating at \SI{100}{\percent}. An estimation of the force from the \textit{i}th thruster can be obtained in closed form with \mbox{$F_i = 1.7 u_i (1 - 0.125 \sum_{j \neq i} u_j )$.} Given the thruster plate's configuration, no more than four thrusters are expected to operate concurrently, as each thruster can be paired with one in the opposite direction.

The three compressed air tanks carry \SI{1.65}{\kg} of compressed air when filled to \SI{300}{\bar}. With a single thruster consuming \SI{3.5}{\gram\per\second} of air, the robot can thus operate a single thruster continuously open for up to \SI{480}{\second}, assuming non-operable below \SI{10}{\bar}. In practice, however, the thrusters are not expected to operate continuously for extended periods and the actual operational time of the robot will depend on the efficiency of the implemented controller.

\begin{figure}[tpb]
   \centering
   \includegraphics[width=1\linewidth, trim={7cm 7cm 7cm 5cm}, clip]{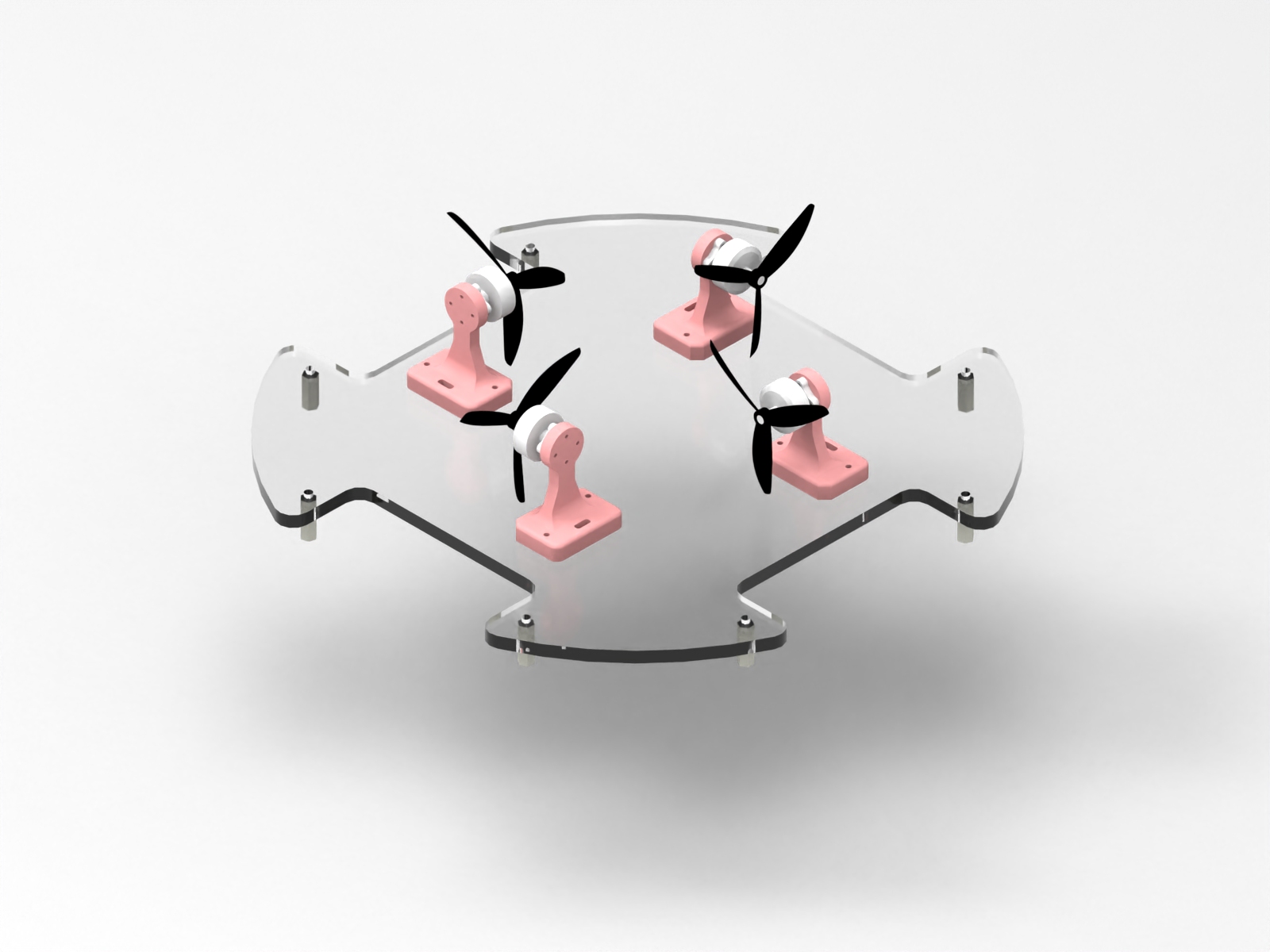}
   \caption{Propeller actuation plate comprising of four motors paired with bi-directional propellers, providing 3 DoF to the platform when integrated. Instead of only controlling the duty cycle of the maximum force, as the thrusters in \cref{fig:thruster_configuration}, the propeller plate will allow scaling of the force magnitude.}
   \label{fig:propeller_plate}
\end{figure}

The propeller plate allows for finer control over the imposed forces on the platform. A preview of the propeller plate is shown in \cref{fig:propeller_plate}. Each motor is paired with a $3.5$ inch propeller and an electronic speed controller (ESC) capable of field-oriented control (FOC). FOC allows us to accurately track the motor rotation independently of the battery level, as long as there is enough power for the requested velocity. Moreover, FOC allows precise rotation speed control at slow speeds, enabling us to inject disturbances into the motion model accurately or replicate orbital dynamics at scale. Each motor is capable of \SI{1.96}{\newton} of thrust in both directions, for a total maximum of \SI{3.92}{\newton} of thrust on each axis and \SI{0.67}{\N\m} of torque.

\begin{figure}[tpb]
   \centering
   \includegraphics[width=1\linewidth]{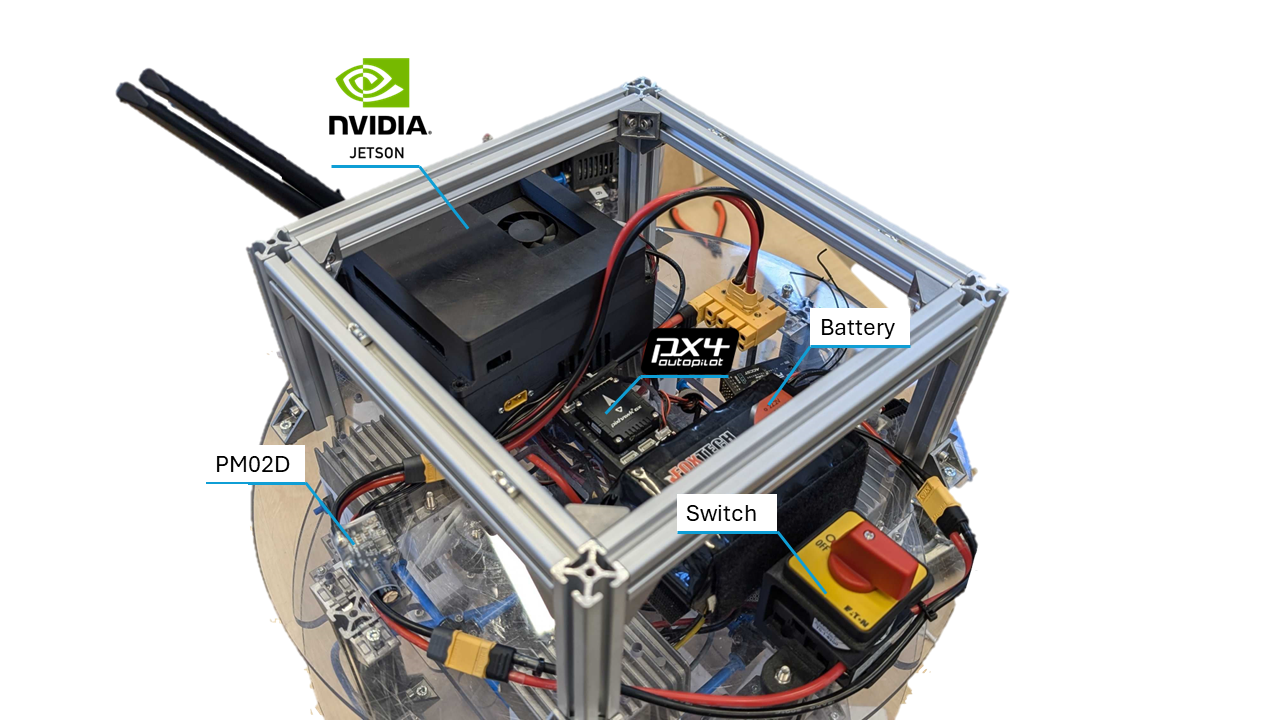}
   \caption{Overview of the avionics layer. A Nvidia Jetson Orin NX is the high-level onboard computer (OBC), while a PX4 6X Mini is the low-level control unit and interface for sensors and actuators. The Jetson OBC runs Ubuntu 22.04, while the PX4 runs NuttX \ac{RTOS}. Also in the image are the PM02D power sensor, the system's battery, and the battery cut-off switch.}
   \label{fig:avionics_overview}
\end{figure}

\subsection{Electronics}

\begin{figure}[tpb]
   \centering
   \includegraphics[width=1\linewidth, trim={3.5cm 3cm 4cm 3cm}, clip]{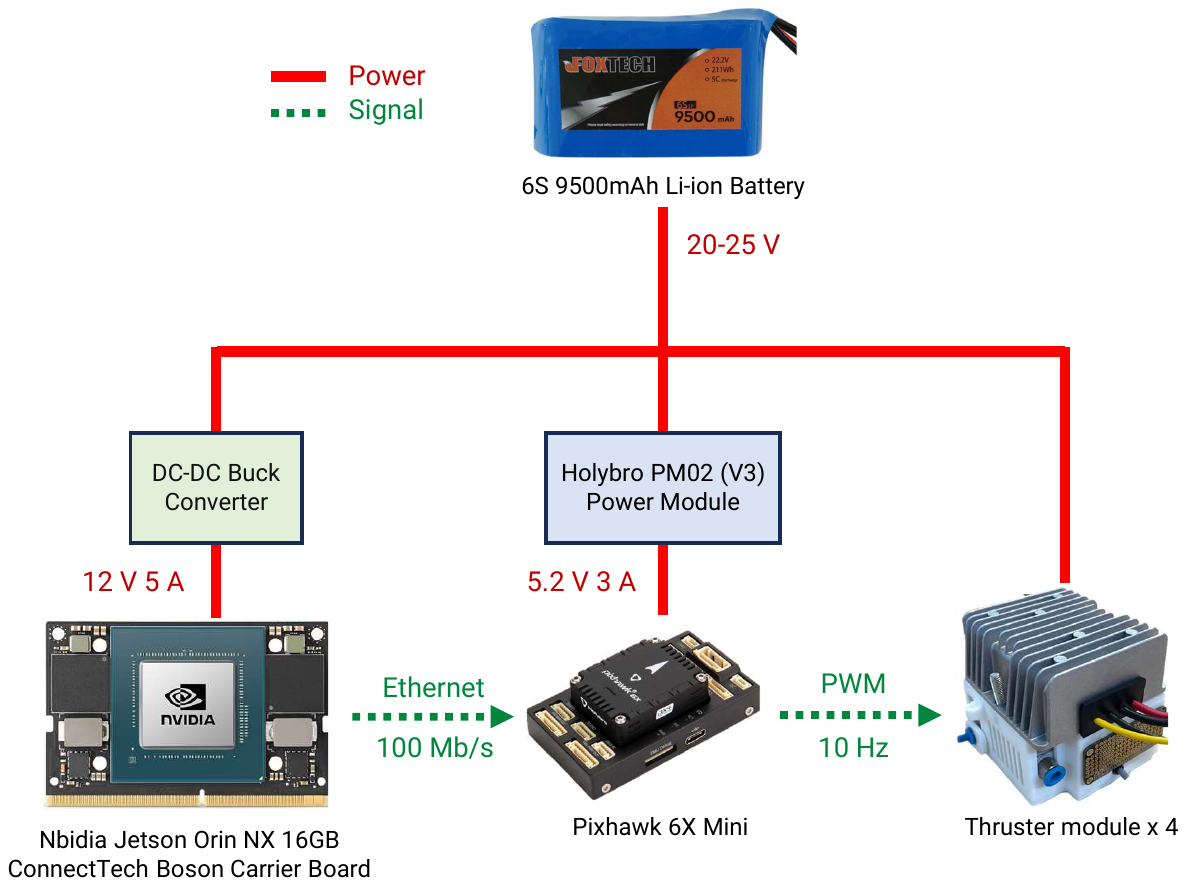}
   \caption{Power and signaling distribution in the avionics subsystem. The power distribution board also provides power to the actuation plate.}
   \label{fig:electronics}
\end{figure}

On top of the actuation plate sits the avionics layer, composed of high-level and low-level computing units, batteries, and a power monitoring module. An overview of this layer is shown in~\cref{fig:avionics_overview}, and the block diagram with the electrical and signals schematic is shown in \cref{fig:electronics}.

We use a Pixhawk 6X Mini as the low-level computing unit running PX4Space as the firmware. This unit comprises triple-redundant and temperature-compensated inertial measurement units (ICM-45686) and two barometers (ICP20100 and BMP388). An ARM Cortex M7 (STM32H753) microcontroller collects the sensor outputs and interfaces with the high-level computer, a Jetson Orin NX, through \SI{100}{\mega\bit\per\second} Ethernet. Pixhawk also serves as an interface with the actuators, capable of driving both thruster and propeller plates simultaneously using, e.g., the two PWM output modules or CAN actuator interfaces. Since this unit runs a \ac{RTOS}, in our case NuttX, it can precisely control the output via CPU interruptions. Lastly, using a separate low-level computing unit provides a safety layer, acting as a failsafe against failures such as stabilization of the platform when the motion capture system odometry estimations are lost, or the high-level controller pushes the system beyond its safety envelope, or the system battery runs low.

The high-level computer is an Nvidia Jetson Orin NX with \SI{16}{\giga\byte} of LPDDR5 RAM, an 8-core ARM Cortex-A78AE with a maximum of \SI{2.2}{\giga\hertz} clock, as well as a GPU composed by 1024 Core Ampere, with 32 Tensor Cores. This unit runs Ubuntu 22.04 as the operating system. The ROS 2 Humble is set up on this platform and interfaces via DDS over Ethernet with PX4 running on the low-level computing unit. This unit also interfaces with the \ac{mocap} over Wi-Fi and passes through the odometry estimations as an external vision system to PX4.

Lastly, the system is powered by a 6-cell \SI{9.5}{\ampere\hour} Lithium Polymer battery monitored with a PM02D digital power module. 

\subsection{Payload Hosting}

\begin{figure}[tpb]
   \centering
   \includegraphics[width=1\linewidth,trim={2cm 8cm 2cm 2cm}, clip]{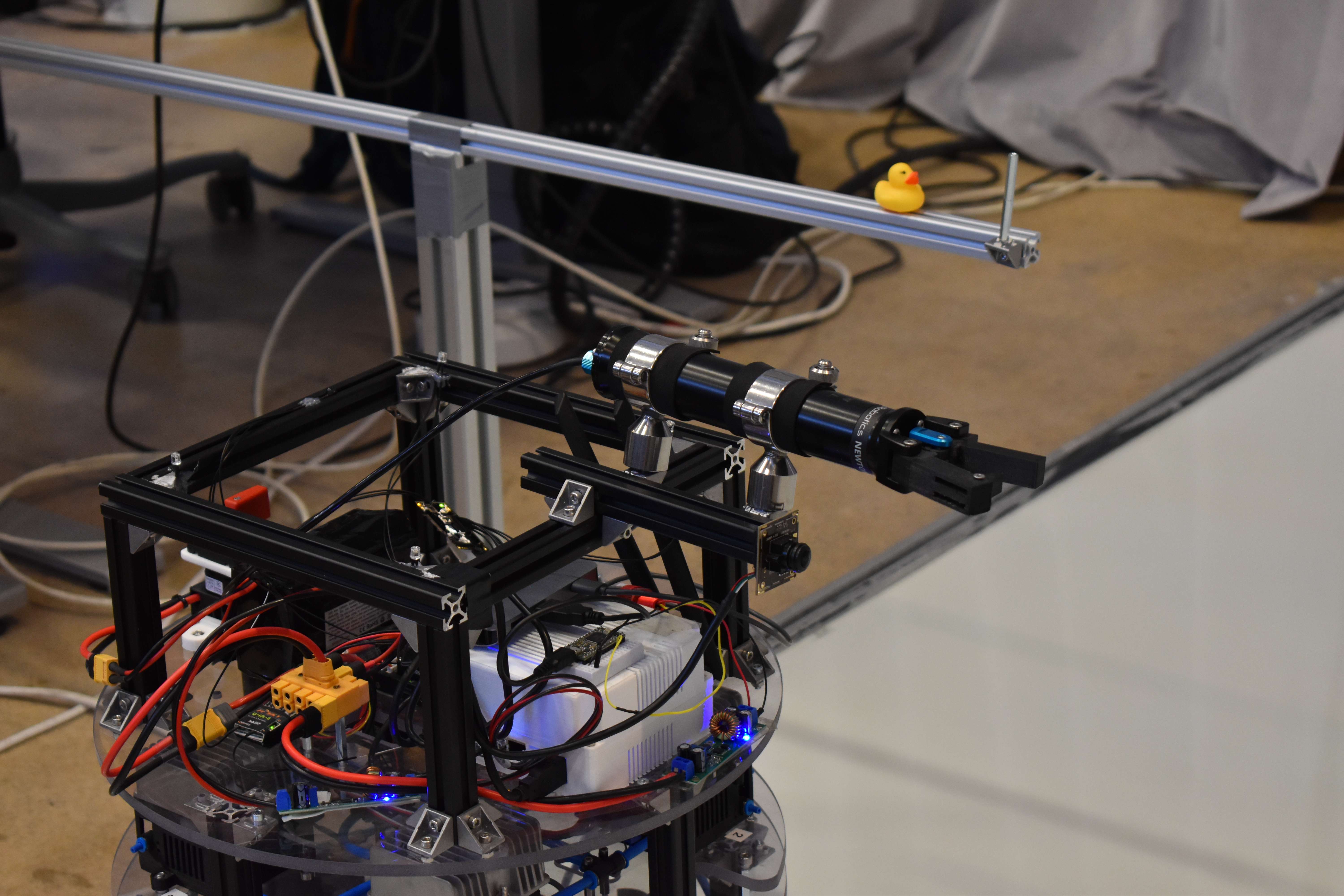}
   \caption{A manipulator attached as a payload to the free-flying platform. The interaction between the manipulator and the onboard avionics is done over USB and a ROS 2 interface node.}
   \label{fig:payload_arm}
\end{figure}

The last layer on the free-flyer is a payload support system. With this architecture, we aim to support other researchers in testing hardware in microgravity conditions and provide a platform for co-development with industry, integrating space-grade hardware into our facilities. \Cref{fig:payload_arm} shows the BlueRobotics Newton Gripper with custom claw, interfacing over USB and ROS 2 with the high-level computer. Another capability is to host actuated platforms - such as CubeSats - while ATMOS generates a trajectory for the system under testing or compensates for the added mass and inertia of the hosting free-flyer. Communication can be done directly with PX4 on the low-level computing unit or through a ROS node on a high-level computer.

\section{Free-flyers Software}
\label{sec:software}
As proposed in this article's introduction, a significant contribution of our facilities is the open-source software stack. In this section, we introduce our software architecture.

\subsection{Software Architecture}

\begin{figure}[tpb]
   \centering
   \includegraphics[width=\linewidth,clip,trim={6.5cm 7.7cm 6.5cm 6.5cm}]{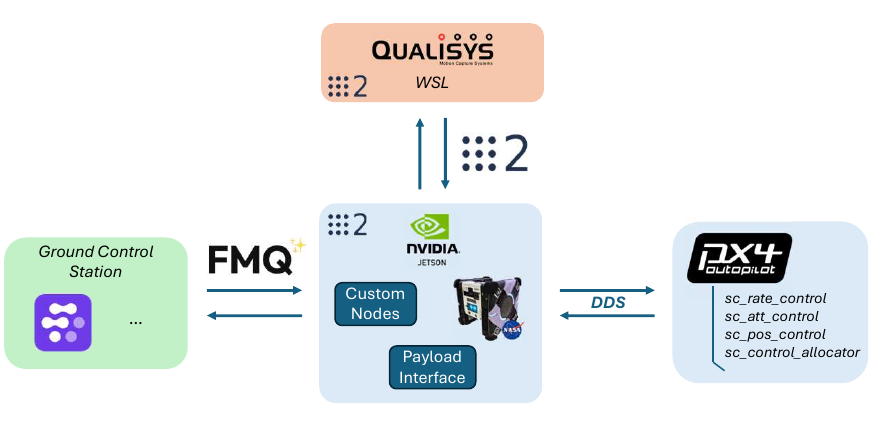}
   \caption{Software architecture of the KTH Space Robotics Laboratory. ROS 2 Humble is the main communication bus between the platforms, while DDS is used to interact with PX4. Lastly, FleetMQ handles the offsite communications with ground control stations or other low-latency applications.}
   \label{fig:software_arch}
\end{figure}

Since multiple computing units are running in real-time and communicating over different protocols, having robust and flexible software solutions is paramount to the efficiency of our facility. In \cref{fig:software_arch}, we shed light on the software architecture in our laboratory facilities.

The software architecture is supported by mainly two communication protocols: Data Distribution Service (DDS), responsible for the interaction between PX4 and the onboard Nvidia Jetson Orin NX, as well as for the ROS 2 Humble middleware; and FleetMQ\footnote{URL: \url{https://www.fleetmq.com/}. Available on 16th February, 2024.}, which provides a low-latency link between the platforms and ground control stations, allowing remote operation of the free-flyers.

\subsection{PX4Space}

\begin{figure*}[!htpb]
\centering
\includegraphics[width=\linewidth,clip,trim={0.5cm 18cm 14cm 11cm}]{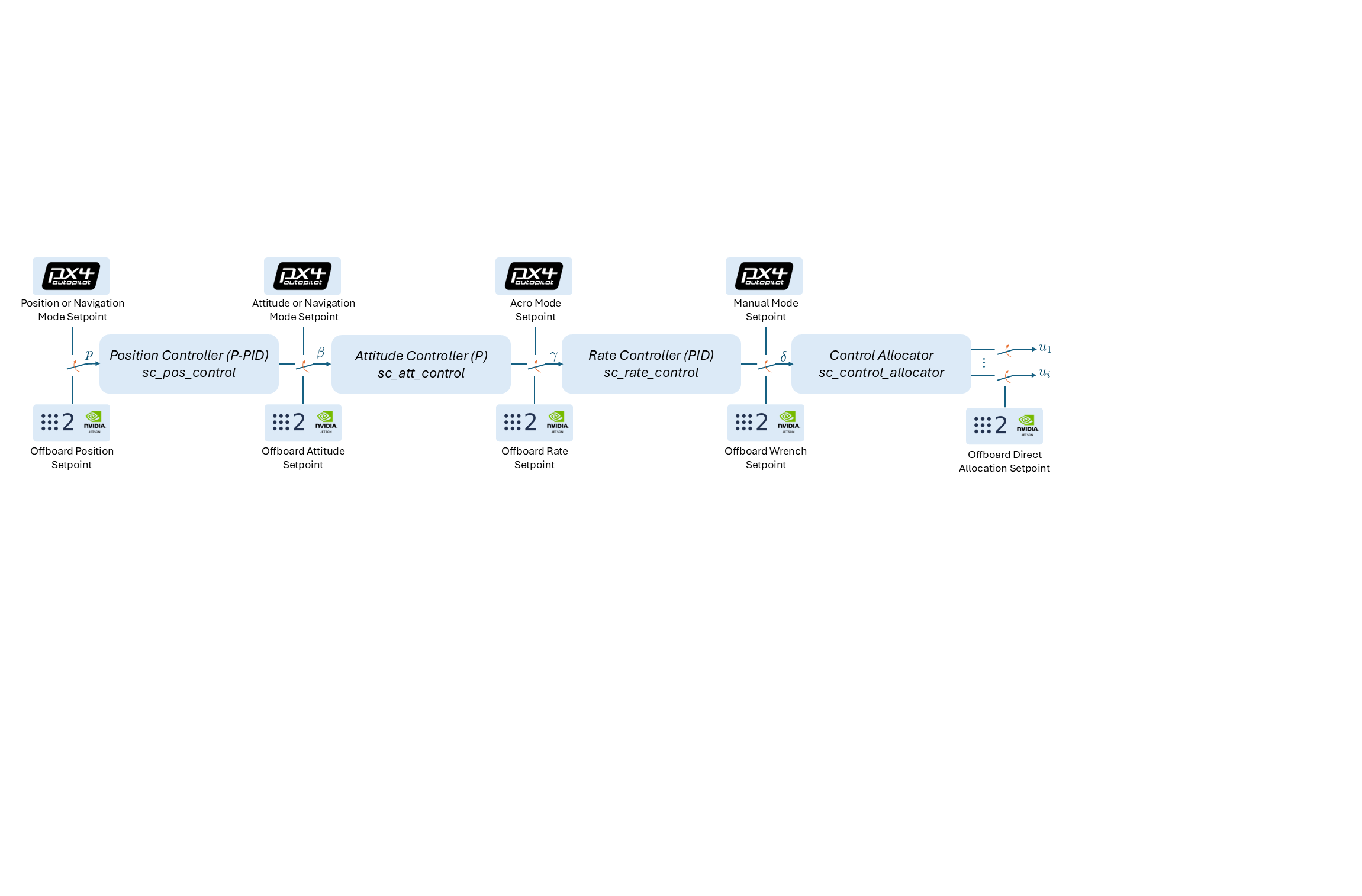}
\caption{Cascaded control scheme implemented in PX4Space. The references $\vecs{p}, \vecs{\beta}, \vecs{\gamma}, \vecs{\delta}$ represent the position, attitude, rate, and thrust/torque (wrench) setpoints, respectively. Only the downstream controllers are active for each setpoint, e.g., an attitude setpoint will disable the position controller, and a wrench setpoint will disable position, attitude, and rate controllers. At the end of the cascade, $\vec{u}_i$ represents the thrust for each $i$th actuator generated by the control allocator or received from offboard.}
\label{fig:px4space_control}
\end{figure*}

PX4Space is a customized version of the open-source PX4-Autopilot\cite{PX4ref} with adapted modules for the spacecraft. These modules provide control functionality for thruster-actuated vehicles. The link to the source code of PX4Space is available on the first page of this article. Alongside this firmware, we also provide an accompanying QGroundControl interface, providing the user easy access to PX4 vehicle setup and parameters, remote control communication, and a PX4 shell terminal.

\begin{remark}
    As of the submission date for this article, parts of PX4Space have been merged with the PX4-Autopilot codebase. We expect the full codebase to be merged no later than May 2025.
\end{remark}

Besides providing interfaces for sensors and actuators, PX4Space also implements the control architecture shown in \cref{fig:px4space_control}. 
The control system comprises three cascaded P and PID controllers. The cascaded loop structure allows PX4Space to be used as a low-level controller of multiple setpoint options and to evaluate different control strategies. The position controller receives a position setpoint $\vec{p}$ and regulates it through a P-controller, generating an internal velocity setpoint tracked by a PID for the velocity error. An attitude setpoint $\vecs{\beta} = \{\vec{f}, \bar{q}\}$ is generated, corresponding to a body-frame thrust $f$ and a quaternion attitude setpoint $\bar{q}$. The position controller can modify the attitude setpoint $\bar{q}$ to ensure feasible tracking for non-holonomic vehicles. The attitude controller implements a P-controller, which generates an angular rate setpoint $\vecs{\gamma} = \{f, \bar{\omega} \}$. Lastly, the rate controller generates a wrench setpoint $\delta = \{ \vec{f}, \vec{\tau}\}$, through a PID to converge the angular velocity to the target.

To actuate each thruster on the platform, the wrench setpoint $\vecs{\delta}$ is processed by the \ac{CA} module. The \ac{CA} allows a selection between a normalized allocation ($\vecs{f}_j \in [0, 1]$) or a metric allocation ($\vecs{f}_j \in [0, f_{max}]$, with $f_{max}$ being the maximum force that each actuator can produce), calculating desired forces for each thruster in Newtons for each thruster $j=1,\dots,8$.
It is worth noting that the above setpoints are specific to the implementation of the cascaded loop structure. These setpoints can be generated from any PX4 modules, manual inputs, or external entities such as an onboard computer. In the PX4Space firmware, the modules \texttt{sc\_pos\_control}, \texttt{sc\_att\_control},  
\texttt{sc\_rate\_control} and \texttt{sc\_control\_allocator} implement each of the modules in \cref{fig:px4space_control} in the same order. 

\subsection{Onboard Computer}

The OBC, at the center of the diagram in \cref{fig:software_arch}, is at the center of all interactions with the free-flyer. The operating system is Ubuntu 22.04, provided by the Nvidia Jetpack 6.0, and running ROS 2 Humble. This unit hosts ROS 2 interfaces for PX4Space and broadcasts selected internal topics in the firmware to ROS 2. Moreover, it communicates via ROS 2 with the MoCap computer to retrieve vehicle's odometry data (pose and velocities), later fused with the PX4 EKF2 estimator. The OBC also serves as a connection endpoint to ground control stations or any other low-latency routes the user requires.

The main task of the OBC is to run all high-level avionics tasks, such as vision-based position estimators, mapping nodes, high-level interface with payloads, or any other task that does not require the strict real-time guarantees of the low-level PX4 system or that cannot be run on its microcontroller. Furthermore, at a later stage, the OBC will allow us to interface with the Astrobee \ac{FSW}, allowing us to test our contributions locally. Since the OBC runs a generic Linux distribution, we also plan to support other \ac{FSW} stacks, providing a modular solution for different software needs.

\subsection{Motion Capture Computer}

The motion capture computer, shown at the top of \cref{fig:software_arch}, runs a Qualisys QTM server that interfaces with the MoCap camera network and ROS 2 clients and publishes high-frequency odometry data. The ROS package responsible for the data translation and augmentation with linear and angular velocities is available in \url{https://github.com/DISCOWER/motion_capture_system}. This package is executed in a Windows Subsystem for Linux (WSL) environment. At this stage, any ROS 2 node can access odometry messages corresponding to the pose and velocities of the free-flyers.

\subsection{Simulation}

\begin{figure}[!ht]
   \centering
   \includegraphics[width=0.8\linewidth]{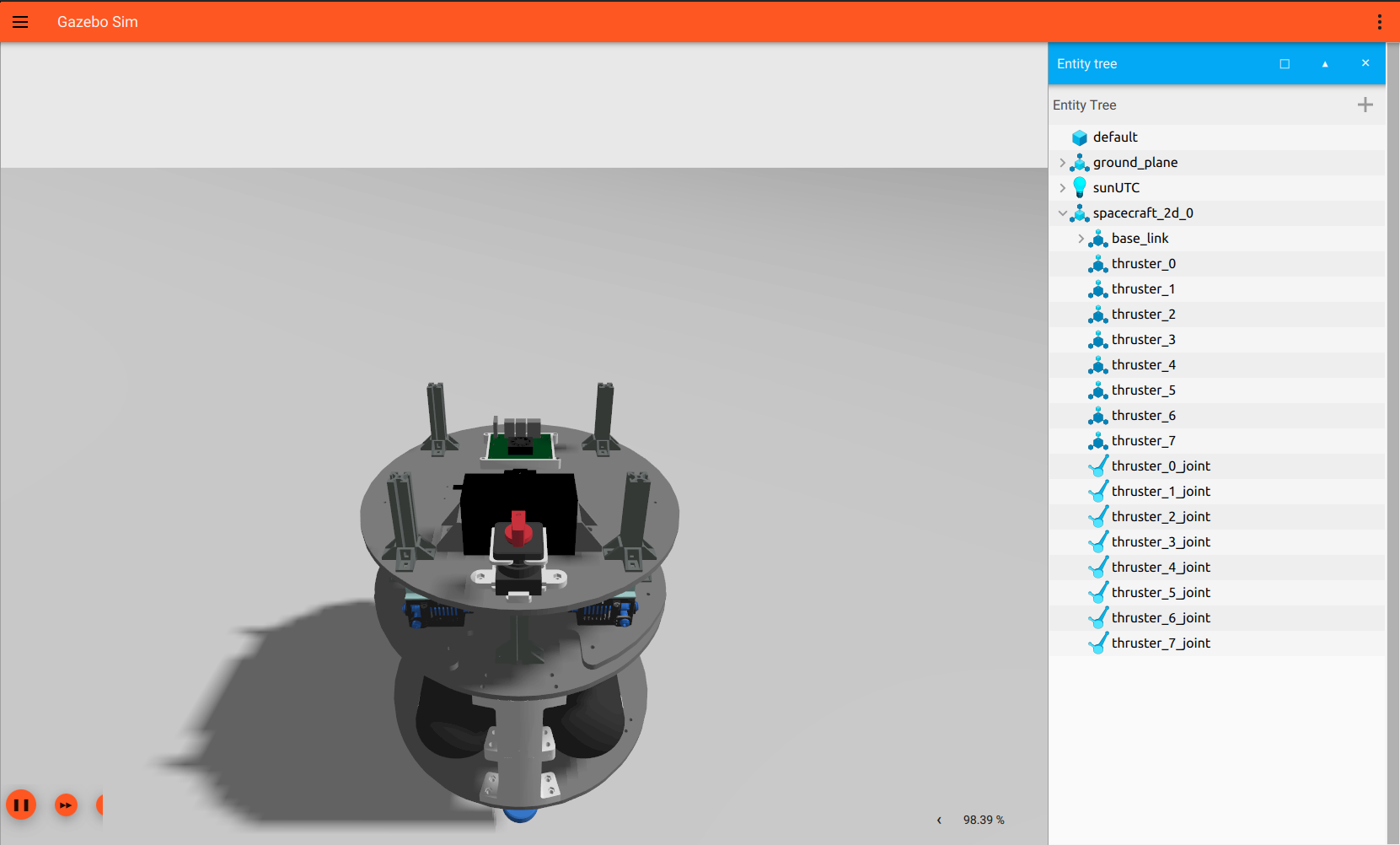}
   \caption{\ac{SITL} simulation in Gazebo. The simulated model interacts with the PX4Space \ac{SITL} setup and allows interaction with any of the control interfaces of the real platform, as well as simulating any of the radio-controlled flight modes.}
   \label{fig:px4_gz}
\end{figure} 

\label{sec:sw_simulation}
The \ac{SITL} simulation environment takes advantage of the Portable Operating System Interface (POSIX) compatibility of PX4, where the firmware can be simulated directly on any POSIX-compatible system. This allows easy and accurate simulation of the deployed software, allowing testing with identical software interfaces to the real system. This environment allows interfacing with all low-level control interfaces such as position, attitude, body rate, body force, and torque, as well as direct input allocation. Furthermore, radio-controlled modes such as manual, acro, or stabilized can be simulated as operating a real system in \ac{SITL}. We currently support Gazebo Garden (and newer versions) for the \ac{SITL} interface through PX4Space. 
It should be noted that the platforms are not restricted to devices that use PX4. As an example, we are currently working towards integrating the NASA Astrobee \ac{FSW} with ATMOS by routing the desired thrusts through PX4Space. Packages for this interfacing and to recreate our laboratory facilities are available in \url{https://github.com/DISCOWER/discower_asim}.

\subsection{Ground Control Station}

\begin{figure}[tpb]
    \centering
    \subfloat[Ground Control Station user interface, built using Foxglove Studio.]{
    \includegraphics[width=0.95\linewidth]{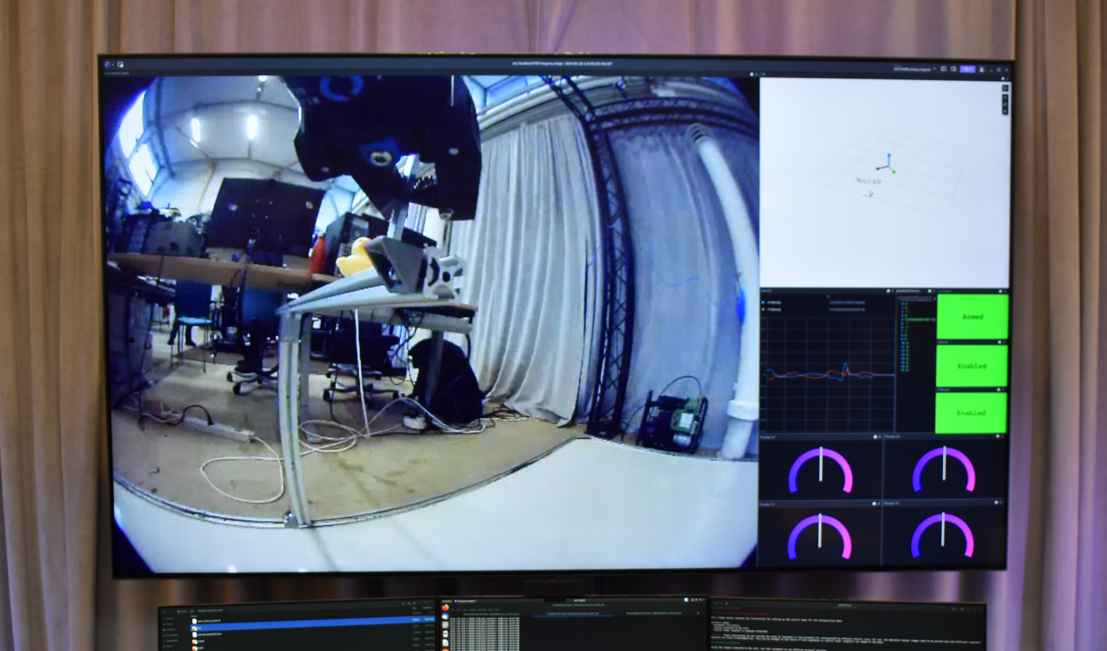}
    \label{fig:gcs_panel_view}
    } \\
    \vspace{2mm}
    \subfloat[A user operates the platform remotely. FleetMQ creates a peer-to-peer connection with minimal latency.]{
    \includegraphics[width=0.95\linewidth,clip,trim={0 0 3cm 3cm}]{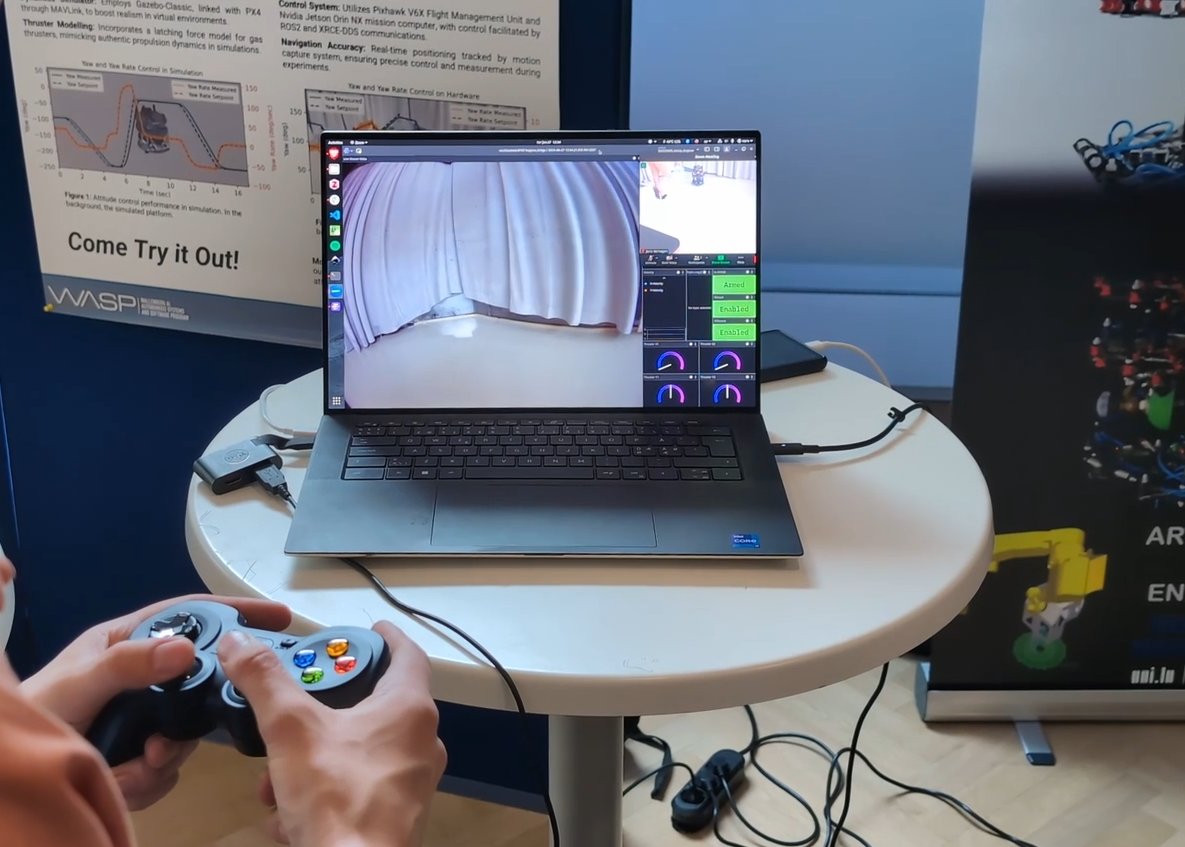}
    \label{fig:gcs_remote_operator}
    } 
    \caption{The ground control station panel provides the teleoperator with a live video feed from the platform, feedback from sensors, and position in the laboratory, and allows remote teleoperation of our platform across large distances. In \cref{fig:gcs_remote_operator}, a platform is operated from Luxembourg, approximately \SI{1300}{\kilo\meter} away.}
    \label{fig:gcs}
\end{figure}

The ground control station (GCS) is a remote operation console for laboratory platforms. Built around Foxglove Studio\footnote{URL: \url{https://foxglove.dev/}. Available on 15th of September, 2024.}, it communicates with each platform over FleetMQ peer-to-peer connection that optimizes the packet routing to the lowest latency path. In the GCS interface, we provide a position of the robotic platform in the MoCap area, feedback from the received thrust commands, vehicle velocity, arming, operation mode, and online status of the free-flyers. Lastly, a low-latency image channel is available, allowing us to remotely operate and navigate the platform. The GCS can be seen in \cref{fig:gcs}.

\section{Autonomy}
\label{sec:autonomy}
In this section, we start by providing an overview of the communication schemes for multi-agent operations at our facility, followed by \ac{NMPC} schemes for nominal and offset-free tracking, and ending with a proposed planner for multi-agent operations. In particular, we consider direct allocation, body force and torque, and body force and attitude rate setpoints, as demonstrated below.

\subsection{Autonomy and Multi-Agent Architecture}

\begin{figure}
    \centering
    \includegraphics[width=0.75\linewidth,trim={0 16cm 36cm 0},clip]{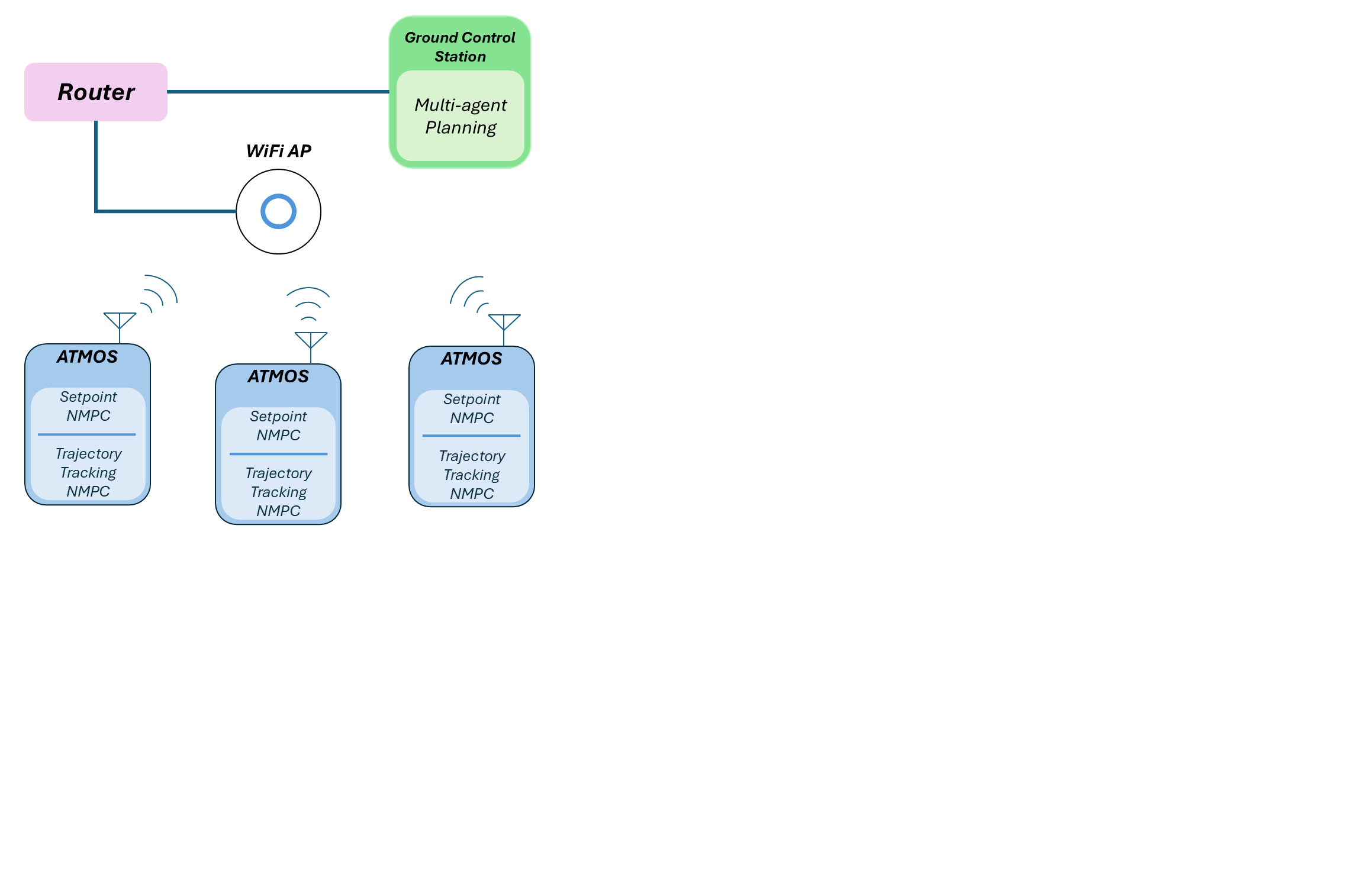}
    \caption{Autonomy architecture for single and multi-agent operations. With three ATMOS platforms available at the facility, single and multi-agent algorithms can be tested. Communication is handled via WiFi. When a multi-agent team is used, a centralized planning module is available on a ground station for coordinating multiple ATMOS units.}
    \label{fig:architecture}
\end{figure}

We first show the laboratory's autonomy architecture, depicted in \cref{fig:architecture}.
The proposed architecture allows us to run multi-agent operations with different communication schemes, from centralized algorithms to distributed control and planning strategies. Agent communication is done over WiFi with a WiFi 7 capable access point. A centralized planning node is also available on the ground control station, which can provide safe trajectories to each agent. Lastly, onboard ATMOS, setpoint stabilization and trajectory tracking \ac{NMPC}s are used to track the planner trajectories to complete the assigned tasks. In the next section, we provide an overview of setpoint stabilization \ac{NMPC} schemes.

\subsection{Nonlinear Model Predictive Control}
\label{ssec:nmpc}
Different control strategies can be tested on ATMOS through the offboard control interface of PX4, either via the onboard high-level compute unit or over a remotely connected client. In this section, three control strategies for setpoint regulation are detailed: i) direct control allocation, ii) body wrench setpoints, and iii) body force and angular rate setpoints. The implemented control schemes are open-source and available at \url{https://github.com/DISCOWER/px4-mpc}.

We choose to employ \iac{NMPC} strategy which minimizes a cost function $J(x,u)$, function of the state $x$ and control input $u$, along a receding horizon of length $N$. The state and control input vectors are constrained to evolve in the polytopes $x\in\mathbb{X}$ and $u\in\mathbb{U}$, and therefore, state and actuation constraints can be taken into account when calculating each control input. At each sampling time $k$ and over each step $n$ of the prediction horizon $N$, a state prediction $x(n+1|k)$ is obtained from the previously predicted (or measured) state $x(n|k)$, the optimized control input $u(n|k)$, and the system dynamics $g(x(n|k), u(n|k))$. The optimization problem results in $N$ predicted states and $N$ control inputs for the system of the form 
$
    \mathbf{x}^*_k = \{ \myvar{x}^*(1|\stepk), \dots, \myvar{x}^*(N|\stepk)\} \label{eq:mpc_pred_state_trajectory} $, and   
    $\mathbf{u}^*_k = \{ \myvar{u}^*(0|\stepk), \dots, \myvar{u}^*(N-1|\stepk) \} \label{eq:mpc_control_trajectory}
$
for a given initial state $\myvar{x}(0|\stepk) = \tildevar{x}(\stepk)$, and an associated optimal cost value $J^*(\tildevar{x}(\stepk))$. Each discrete control input is applied to the system in a Zero Order Hold (ZOH) fashion - a piece-wise constant input between sampling instances, that is, $\myvar{u}(t) = \myvar{u}^*(\stepk) \ \forall t \in [k\Delta t, (k+1)\Delta t)$ - abbreviated to $t \in [k, k+1)$ in this manuscript. 

The \ac{NMPC} optimization problem is then defined as
\begin{subequations}
\label{eq:MPC_formulation}
\begin{align}
J^*( \tildevar{x}(& \stepk)) = \min_{\mathbf{u}_k}  \  J(\myvar{x}(n| \stepk), \myvar{u} (n|\stepk)) \\
 \text{s.t.:} \hspace{2mm} & \myvar{x}(i+1| \stepk) = g(\myvar{x}(i| \stepk), \myvar{u}(i| \stepk)) \\
 & \myvar{u}(i|\stepk) \in \mathbb{U}, \ \forall i\in\mathbb{N}_{[0,N-1]} \label{eq:mpc_set_cond3} \\
 & \myvar{x}(i|\stepk)  \in \mathbb{X}, \  \forall n\in\mathbb{N}_{[0,N]} \label{eq:mpc_set_cond1}\\
 & \myvar{x}(N|\stepk) \in \mathbb{X}_N \subset \mathbb{X} \label{eq:mpc_set_cond2}\\
 & \myvar{x}(0|\stepk) = \myvar{x}(\stepk)
\end{align}
\end{subequations}
where the cost function for setpoint stabilization is given by
\begin{subequations}
\label{eq:centralized_mpc}
\begin{align}
    J(\bar{\vec{x}},\vec{x},\vec{u}) &= \sum_{n=0}^{N-1} l\big(\bar{\vec{x}}(n|k),\vec{x}(n|k),\vec{u}(n|k) \big) \nonumber \\
    & \qquad + V\big(\bar{\vec{x}}(n|k),\vec{x}(n|k)\big), \\
    l\big( \bar{\vec{x}},\vec{x}, \vec{u}\big) &= \|\vec{e}(n|k)\|^2 _{\mat{Q}} + \| \vec{u}(n|k) \|^2_\mat{R}, \\
    V\big(\bar{\vec{x}},\vec{x}\big) &= \| \vec{e}(N|k) \|^2_{\mat{Q}_N} 
\end{align}
\end{subequations}
where $e$ is the error of the state $x$ with respect to the reference $\bar{x}$. The set $\mathbb{X}_N$ is a terminal control invariant set under a static state feedback controller, such as $\myvar{u}_K(t) = K \myvar{x}(t)$, for a given gain matrix $K$. It is typical \cite{chen1998quasi} to use Linear Quadratic Regulators (LQR) and associated control invariant sets as terminal sets in \ac{NMPC}.

Depending on the chosen control model - direct control allocation, body force and torque, or body force and angular rate - different models for $g(\cdot)$, state $x$ and error $e$ are used, and correspondingly, different state and control constraint sets. Below, we explain the models used in each scenario.

\subsubsection{Direct Control Allocation} In direct control allocation, each thruster is modeled in the dynamics through allocation matrices $D$ and $L$. These represent the resulting center-of-mass force and torque applied by each thruster. The state is $x=[p,v,q,\omega]\in\mathbb{X}\subset\mathbb{R}^9\times\mathbb{SO}(3)$,$p\in\mathbb{R}^3$ represents the cartesian position, $v\in\mathbb{R}^3$ the linear velocity, $q\in\mathbb{SO}^3$ the attitude quaternion, $\omega\in\mathbb{R}^3$ the angular velocity, and the control signal $u=[u_1,...,u_8]\in\mathbb{U}\subset\mathbb{R}^8$ is the desired thrust for each thruster. In this case, $g(x,u)$ is the discretized (using, for instance, a 4th order Runge Kutta method) version of the model
\begin{subequations}
\label{ctl:da}
\begin{align}
    \dot{p} &= v, \\
    \dot{v} &= \frac{1}{m}R(q)^T D u, \\
    \dot{q} &= \frac{1}{2}\Xi(q)\omega, \\
    \dot{\omega} &= M^{-1}(L u + \omega \times M \omega), 
\end{align}
\end{subequations}
where $D\in\mathbb{R}^{3\times 4}$ and $L\in\mathbb{R}^{3\times 4}$ allowing us to use eight thrusters with four decision variables, $R(q)\in\mathbb{SO}^3$ \cite[Eq. 76]{trawny2005indirect} represents the rotation matrix associated with the quaternion $q$, $\Xi(q)$ \cite[Eq. 54]{trawny2005indirect} represents the skew-symmetric matrix of $q$, $m\in\mathbb{R}_{>0}$ represents the system mass, and $M\in\mathbb{R}^{3\times3}$ its inertia matrix. Similarly, the constraint set $\mathbb{U}$ represents the minimum and maximum bounds for each thruster, as $\mathbb{U} \triangleq \{ u_{[j]}\in\mathbb{R} | -f_{max} \leq u_{[j]} \leq f_{max}, j=1,...,4\}$, and where each $u_{[j]}$ corresponds to an aligned thruster pair ($u_{[j]} > 0$ actuates one thruster, whereas $u_{[j]} < 0$ actuates the opposite one). The error $e$ for this strategy is defined as
\begin{align}
    \vec{e}(n|k) &= \begin{cases}
      \bar{\vec{x}}(n|k) - \vec{x}(n|k), \vec{x}\leftarrow \{\vec{p}, \vec{v}, \vecs{\omega}\}\\
      1 - (\bar{\vec{x}}(n|k)^T \vec{x}(n|k))^2, \vec{x} \leftarrow  \{ \vec{q} \} \end{cases} \label{eq:full_error}.
\end{align}

\subsubsection{Body Force and Torque} In this scenario, the \ac{NMPC} generates desired body-frame force and torques. The model's continuous-time version of $g(x,u)$ is similar to the one considered for control allocation and is defined as
\begin{subequations}
\label{ctl:wrench}
\begin{align}
    \dot{p} &= v, \\
    \dot{v} &= \frac{1}{m}R(q)^T f, \\
    \dot{q} &= \frac{1}{2}\Xi(q)\omega, \\
    \dot{\omega} &= M^{-1}(\tau + \omega \times M \omega), 
\end{align}
\end{subequations}
where the control input vector $u$ is defined as $u=[f^T,\tau^T]\in\mathbb{U}\subset\mathbb{R}^6$ and the state as $x=[p,v,q,\omega]\in\mathbb{X}\subset\mathbb{R}^9\times\mathbb{SO}(3)$, similarly to the Direct Control Allocation model. An important distinction between controlling a body force and torque and the control allocation model is the necessarily more conservative control constraint bounds to avoid saturation. However, in certain scenarios, using a body wrench input may use fewer control variables, leading to a faster control rate. The control constraint set is defined as 
$\mathbb{U} \triangleq \{ f\in\mathbb{R}^3, \tau\in\mathbb{R}^3 | -f_{max} \leq f_{[x,y,z]} \leq f_{max}, -\tau_{max} \leq \tau_{[x,y,z]} \leq \tau_{max}\}$ and the setpoint $\delta$ defined as $\delta = [f^T,\tau^T]$. In this case, the error $e$ is defined in the same manner as in \cref{eq:full_error}.

\subsubsection{Body Force and Angular Rate} Lastly, we tested offboard control with a body force and desired angular rate. This simpler model considers as control inputs a target force and angular velocity, state as $x=[p,v,q]\in\mathbb{X}\subset\mathbb{R}^6\times\mathbb{SO}(3)$, and $g(x,u)$ the discretized version of
\begin{subequations}
\label{ctl:rate}
\begin{align}
    \dot{p} &= v, \\
    \dot{v} &= \frac{1}{m}R(q)^T f, \\
    \dot{q} &= \frac{1}{2}\Xi(q)\omega.
\end{align}
\end{subequations}
The control constraint set is defined as $\mathbb{U} \triangleq \{ f\in\mathbb{R}^3, \omega\in\mathbb{R}^3 | -f_{max} \leq f_{[x,y,z]} \leq f_{max}, -\omega_{max} \leq \omega_{[x,y,z]} \leq \omega_{max}\}$ and the setpoint $\gamma =  [f^T,\omega^T]$.
Furthermore, note that the attitude dynamics are not considered in this model, resulting in a suboptimal controller when compared to the previous scenario, with the advantage of being faster to solve online. The error $e$ for this control approach is given by
\begin{align}
    \vec{e}(n|k) &= \begin{cases}
      \bar{\vec{x}}(n|k) - \vec{x}(n|k), \vec{x}\leftarrow \{\vec{p}, \vec{v}\}\\
      1 - (\bar{\vec{x}}(n|k)^T \vec{x}(n|k))^2, \vec{x} \leftarrow  \{ \vec{q} \} \end{cases}.
\end{align}

\subsection{Offset-Free Nonlinear Model Predictive Control}
\label{ssec:of_nmpc}

The dynamics for the MPC controller proposed in \cref{ssec:nmpc} assume that the system is undisturbed. However, often, this is not the case. On microgravity testbeds, uneven surfaces and model imperfections might cause a bias in the system dynamics, inducing steady-state errors on reference tracking tasks. To overcome this limitation, we propose using an offset-free NMPC scheme based on the work in \cite{morari2012nonlinear}. Let us consider the body force and torque model in \cref{ctl:wrench}, modified to include the disturbances $d_v\in\mathbb{R}^3$ and $d_\omega\in\mathbb{R}^3$ in the translation and attitude dynamics, obtaining the disturbed model
\begin{subequations}
\label{ctl:disturbance_wrench}
\begin{align}
    \dot{p} &= v, \\
    \dot{v} &= \frac{1}{m}R(q)^T f + d_v, \\
    \dot{\vv{q}} &= \frac{1}{2}\tilde{\Xi}(\vv{q})\omega, \label{subeq:q_lin} \\
    \dot{\omega} &= M^{-1}(\tau + \omega \times M \omega) + d_\omega. 
\end{align}
\end{subequations}
As the disturbances are unknown, we employ an \ac{EKF} to estimate $\hat{d}_v$ and $\hat{d}_\omega$ using the residual of the expected dynamics with respect to $d_\omega = d_v = \begin{bmatrix} 0 & 0 & 0\end{bmatrix}^T$ and the true dynamics in \cref{ctl:disturbance_wrench} in the \ac{EKF} measurement update step, at each sampling time $k$. As the system in \cref{ctl:disturbance_wrench} is not fully controllable, we linearize the quaternion dynamics with the model in \cref{subeq:q_lin} where $\vv{q}$ corresponds to the vector component of the quaternion $q$, defined as $q := \begin{bmatrix} q_w & \vv{q} \end{bmatrix}^T$. Note that the rotation matrix $R(q)$ is still fully defined, as the scalar component of $q$ can be obtained with $q_w = \sqrt{1 - \vv{q}^T \vv{q}}$, and the matrix $\tilde{\Xi}$ corresponds to the three last rows of $\Xi$. At each sampling time, we solve the \ac{NMPC} in \cref{eq:MPC_formulation} with $\hat{x}(i+1| \stepk) = g(\hat{x}(i| \stepk), \myvar{u}(i| \stepk))$ given by the discretization of \cref{ctl:disturbance_wrench}, obtaining $u^*(k)$. The estimates $\hat{d}_v$ and $\hat{d}_\omega$ are obtained from the residuals of the estimated $\hat{x} = [\hat{p},\hat{v},\hat{\vv{q}},\hat{\omega}, \hat{d}_v,\hat{d}_\omega]\in\mathbb{X}\subset\mathbb{R}^{15}\times\mathbb{SO}(3)$ and measured $x=[p,v,\vv{q},\omega]$ states, and optimal control input $u^*(k)$.

It is worth noting that the wrench model in \cref{ctl:disturbance_wrench} can be extended to the direct allocation model of \cref{ctl:da}. Translation disturbances can also be included in \cref{ctl:rate}, but this simplified model does not allow disturbances in the attitude dynamics.

\subsection{Thruster Force to Pulse-Width Modulation}
As each actuator requires a \ac{PWM} signal as an input, we must convert our desired actuator thrust to a \ac{PWM} duty cycle. Consider the variable $\lambda\in[0, 1]$, where $\lambda=1$ corresponds to a thruster $j$ being open for the entire time duration between sampling times, $\lambda=0$ being closed for the same duration, and $\lambda\in(0,1)$ the thruster being open a corresponding percentage of time between time sampling times. Then, the duty-cycle $\lambda$ for a given thruster maximum force $\bar{f}$ is given by:
\begin{equation*}
     \lambda = \frac{\vec{f}_{i}}{f_{max}}, j=1,\dots,8.
\end{equation*}

\ifdefined\FinalVersion
\subsection{Planning Schemes}
\label{ssec:planning_schemes}
While MPC provides real-time control by tracking trajectories and rejecting disturbances, high-level planning frameworks play a crucial role in guiding autonomous systems, especially in complex, multi-agent scenarios. To this end, we here present a lightweight, adaptable planning scheme for multi-robot motion planning with Signal Temporal Logic~\cite{maler_monitoring_2004} (STL) specifications.
The STL planner considers specifications over real-valued signals in both the signal dimension and the time dimension, making it a powerful tool for specifying properties of dynamical systems. Examples include, but are not limited to, fuel consumption, keeping vehicles within certain regions for a given amount of time, avoiding obstacles, and triggering actions at specific time instances. An important aspect of STL is the inherent validation of the proposed plan, in terms of feasibility and correctness.

We define a simplified fragment of STL that specifies desired properties of $n$-dimensional, finite, continuous-time signals $\mathbf{x}:\mathbb{R}_{\ge 0} \rightarrow X \subseteq \mathbb{R}^n$.

\begin{definition}[Fragment of Signal Temporal Logic]
\label{def:stl}
    Let bounded time intervals $I$ be in the form $[t_1,t_2]$, where, for all, $I \subset \mathbb{R}_{\ge 0}$, $t_1,t_2 \in \mathbb{R}_{\ge 0}, t_1 \leq t_2$. 
    Let $\mu:X \rightarrow \mathbb{R}$ be a linear real-valued function, and let $\mathrm{pred}:X \rightarrow \mathbb{B}$ be a linear predicate defined according to the relation $\mathrm{pred}(\mathbf{x}) := \mu(\mathbf{x}) \ge 0$. The set of predicates is denoted $\textit{AP}$. 
    We consider a fragment of STL, recursively defined as
    \begin{equation*}
    \label{eq:stl_fragment}
        \begin{aligned}
            \psi &::= \mathrm{pred} \mid \neg \mathrm{pred} \mid \diamondsuit_{I}\mathrm{pred} \mid \Box_{I} \mathrm{pred} \\
            \phi &::= \psi \mid \psi_1 \land \psi_2 \mid \psi_1 \lor \psi_2 
        \end{aligned}
    \end{equation*}
    where $\mathrm{pred} \in \textit{AP}$. 
    The symbols $\wedge$ and $\lor$ denote the Boolean operators for conjunction and disjunction, respectively; and $\diamondsuit_{I}$ and $\Box_{I}$ denote the temporal operator {\em Eventually} and {\em Always}.
\end{definition}
In this fragment, we consider the bounded {\em Always} and {\em Eventually} operators, (requiring to have a predicate hold for all time $t \in I$ and for any time $t \in I$, respectively) and allow conjunctions and disjunctions of any combinations of these. 
Predicates could then consider the position of spacecraft w.r.t an area of interest, its propellant or battery levels, solar panel illumination, separation distance to debris, or radiation exposure.

The consideration of specifications over real-valued signals allows us to define STL robustness metrics maximizing the numeric values of the predicates via robustness semantics.
STL robustness metrics could maximize the satisfaction of the real-valued predicate functions w.r.t. the Boolean and temporal operators in the formula $\phi$.
\emph{Spatial robustness} is then a quantitative way to evaluate satisfaction or violation of a formula:
\begin{subequations}
\label{eq:stl_space_robustness}
\begin{align*}
    \rho_{\mathrm{pred}}(t,x) &= \mu(x(t)) \\
    \rho_{\neg \mathrm{pred}}(t,x) &= -\rho_{\mathrm{pred}}(t,x) \\
    \rho_{\diamondsuit_{I} \mathrm{pred}}(x) &= \max_{\tau \in t+I}(\rho_{\mathrm{pred}}(\tau,x)) \\
    \rho_{\Box_{I} \mathrm{pred}}(x) &= \min_{\tau \in t+I}(\rho_{\mathrm{pred}}(\tau,x)) \\
    \rho_{\phi_1 \land \phi_2}(x) &= \min(\rho_{\phi_1}(x),\rho_{\phi_2}(x)) \\
    \rho_{\phi_1 \lor \phi_2}(x) &= \max(\rho_{\phi_1}(x),\rho_{\phi_2}(x))
\end{align*}
\end{subequations}
Intuitively, this can be understood that for the {\em Eventually} operator, we may look at the best-case satisfaction in the interval $I$ as it only needs to satisfy the predicate once anyway. For the {\em Always} operator, we look at the worst-case satisfaction in the interval $I$ as this is the most critical point that would violate the operator under the minimal spatial disturbance.

Although $x$ may be any measurable signal of the system, we now focus on $x$ being the state or trajectory of the robot.
In general, using the trajectory parametrization of $x$ as in the MPC controllers will lead to a problem that is too complex to solve in a reasonable time. 
Instead, we parametrize the trajectory of the robot using temporal and spatial B\'ezier curves, $h(s)$ and $r(s)$ respectively. 
These B\'ezier curves are then coupled to parametrize the position $x$ according to $r(s) := x(h(s))$ and the velocity $\dot{x}$ according to its derivative $\dot{r}(s) := \dot{x}(h(s))\dot{h}(s)$
For details, we refer to \cite{verhagen2024temporally}. 
This parametrization allows for an entire trajectory, $\mathbf{x}^b$, to be built of concatenated segments $x_i, i \in \{1,\ldots,N\}$; $\mathbf{x}^b = \{x_1,\ldots,x_N\}$. Each segment $i$ is then parametrized by $h_i(s)$ and $r_i(s)$, spanning an arbitrarily short or long duration and distance respectively.
We formulate the following Mixed-Integer Linear Problem
\begin{align*}
\argmax_{h_i,r_i, \forall i \in \{1,\ldots,N\}} \quad & \rho_{\phi}(\mathbf{x}^b) \\
\textrm{s.t.} \quad & \mathbf{x}^b(t_0) = x_{t_0}, \mathbf{x}^b(t_f) = x_{t_f}, \\
                    & \mathbf{x}^b(t) \in \mathcal{W}_{\mathit{free}}, \forall t \in [t_0,t_f],\\ 
                    & \dot{\mathbf{x}}^b(t) \in \mathcal{V}, \forall t \in [t_0,t_f],.
\end{align*}
This problem maximizes the spatial robustness while ensuring the initial- and final state are adhered to, the robot is in the workspace, and velocity constraints are adhered to.
In addition, we add a cost term that penalizes the acceleration of the plan \cite{verhagen2024temporally}, promoting smoothness.

\fi

\section{Results}
\label{sec:prel_results}
\begin{figure*}[tpb]
    \centering
    \begin{subfigure}[t]{0.45\linewidth}
        \centering
        \includegraphics[width=\linewidth]{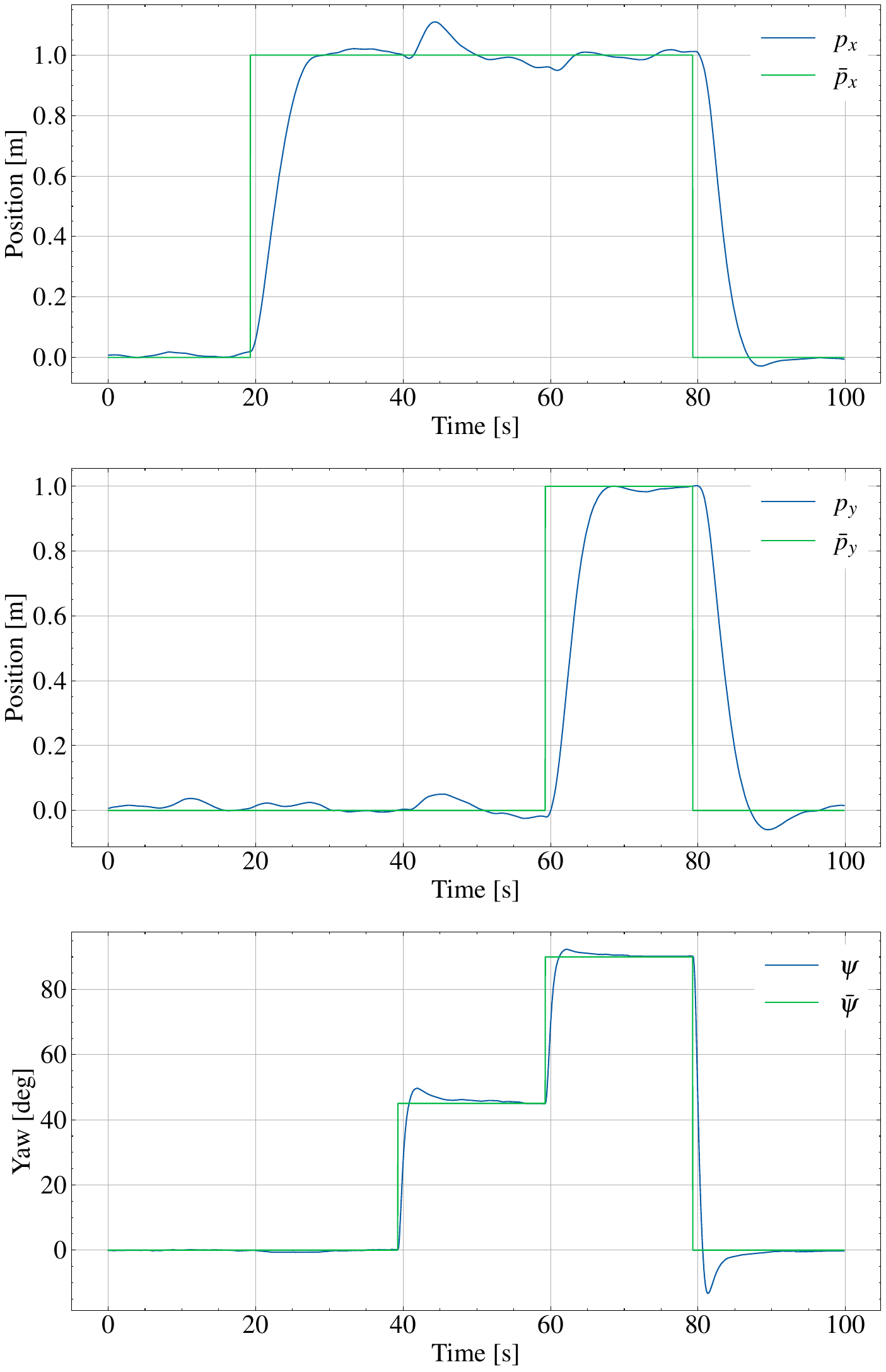}
        \caption{ATMOS \ac{SITL} performance with direct control allocation.} \label{subfig:sitl_da_states}
    \end{subfigure} 
    \hspace{2mm}
    \begin{subfigure}[t]{0.45\linewidth}
        \centering
        \includegraphics[width=\linewidth]{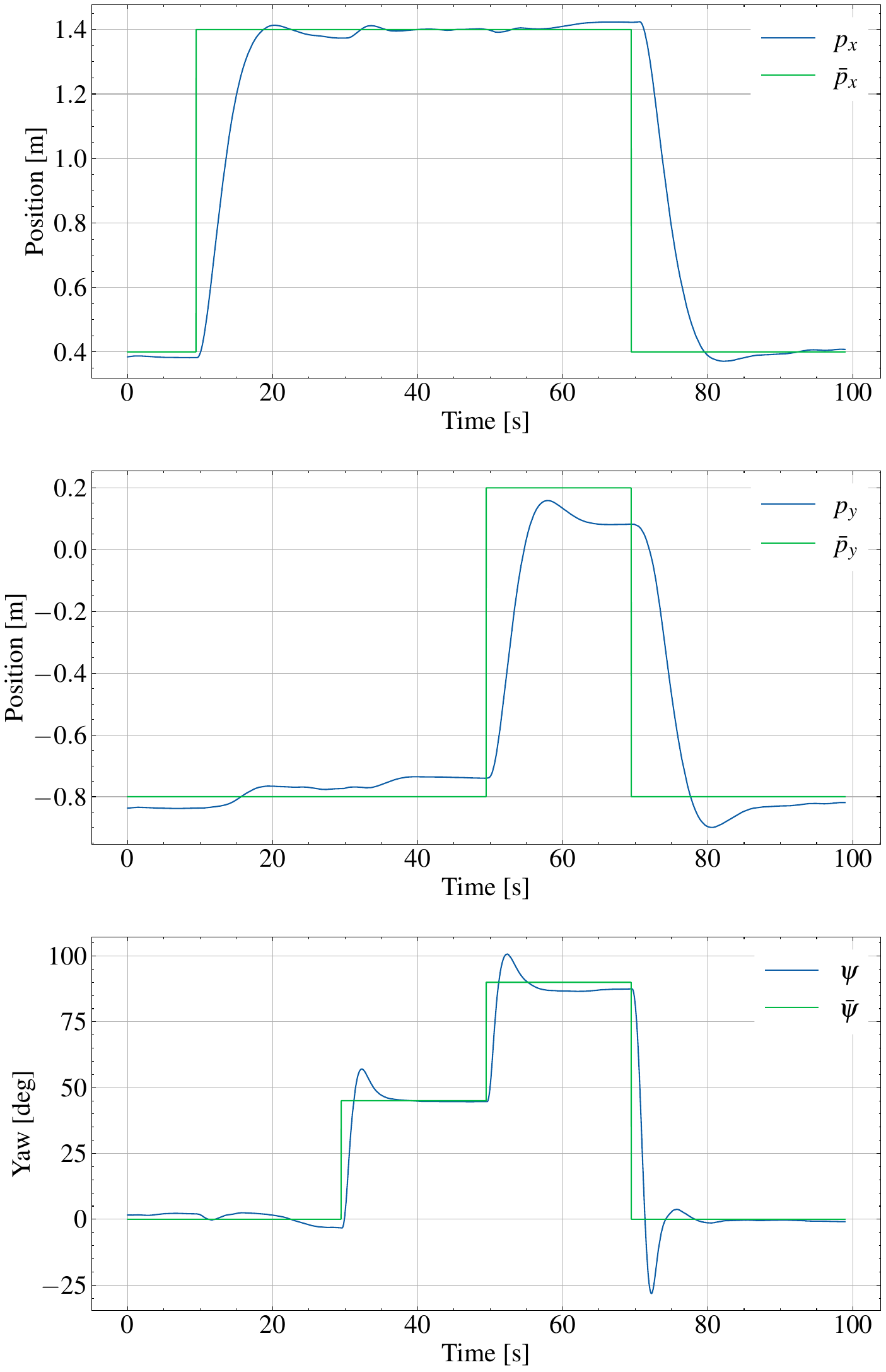}
        \caption{ATMOS hardware performance with direct control allocation.} \label{subfig:real_da_states}
    \end{subfigure}
    \\
    \vspace{2mm}
    \begin{subfigure}[t]{0.45\linewidth}
        \centering
        \includegraphics[width=\linewidth]{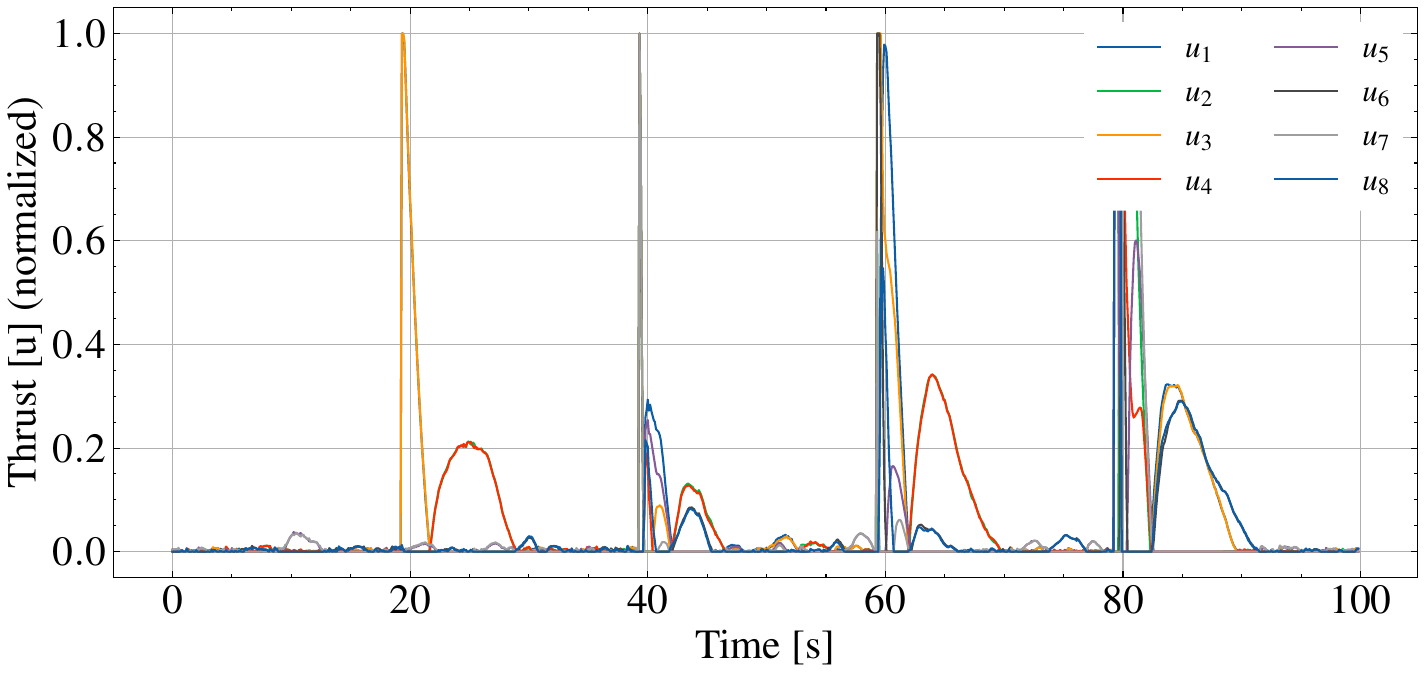}
        \caption{ATMOS \ac{SITL} control inputs with direct control allocation.} \label{subfig:sitl_da_inputs}
    \end{subfigure}
    \hspace{2mm}
    \begin{subfigure}[t]{0.45\linewidth}
        \centering
        \includegraphics[width=\linewidth]{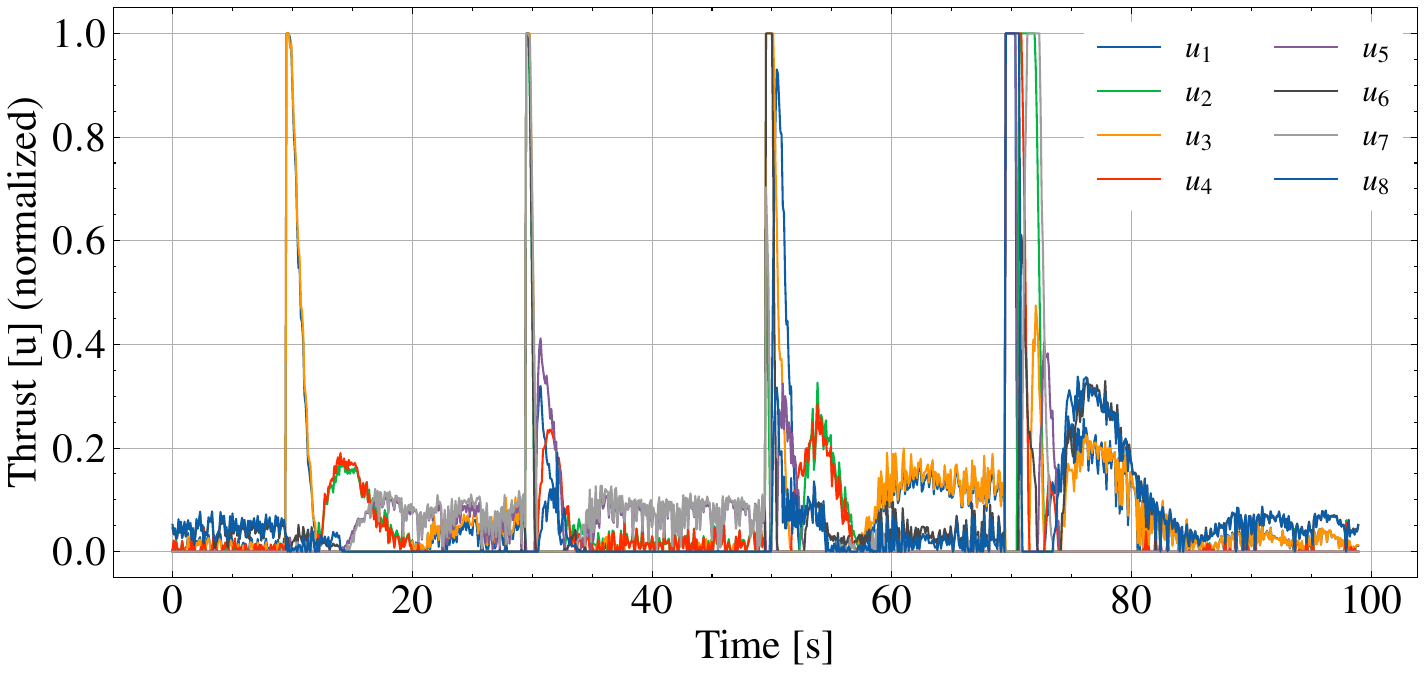}
        \caption{ATMOS hardware control inputs with direct control allocation.} \label{subfig:real_da_inputs}
    \end{subfigure}
    \caption{ATMOS \ac{SITL} and hardware performance when using the \ac{NMPC} scheme in \cref{eq:centralized_mpc} and considering the Direct Control Allocation model in \cref{ctl:da}. Despite the hardware steady-state error, we can observe that the \ac{SITL} transients closely follow the experimental results, validating the \ac{SITL} dynamics model.}
    \label{fig:direct_allocation}
\end{figure*}

\begin{figure*}[tpb]
    \centering
    \begin{subfigure}[t]{0.45\linewidth}
        \centering
        \includegraphics[width=\linewidth]{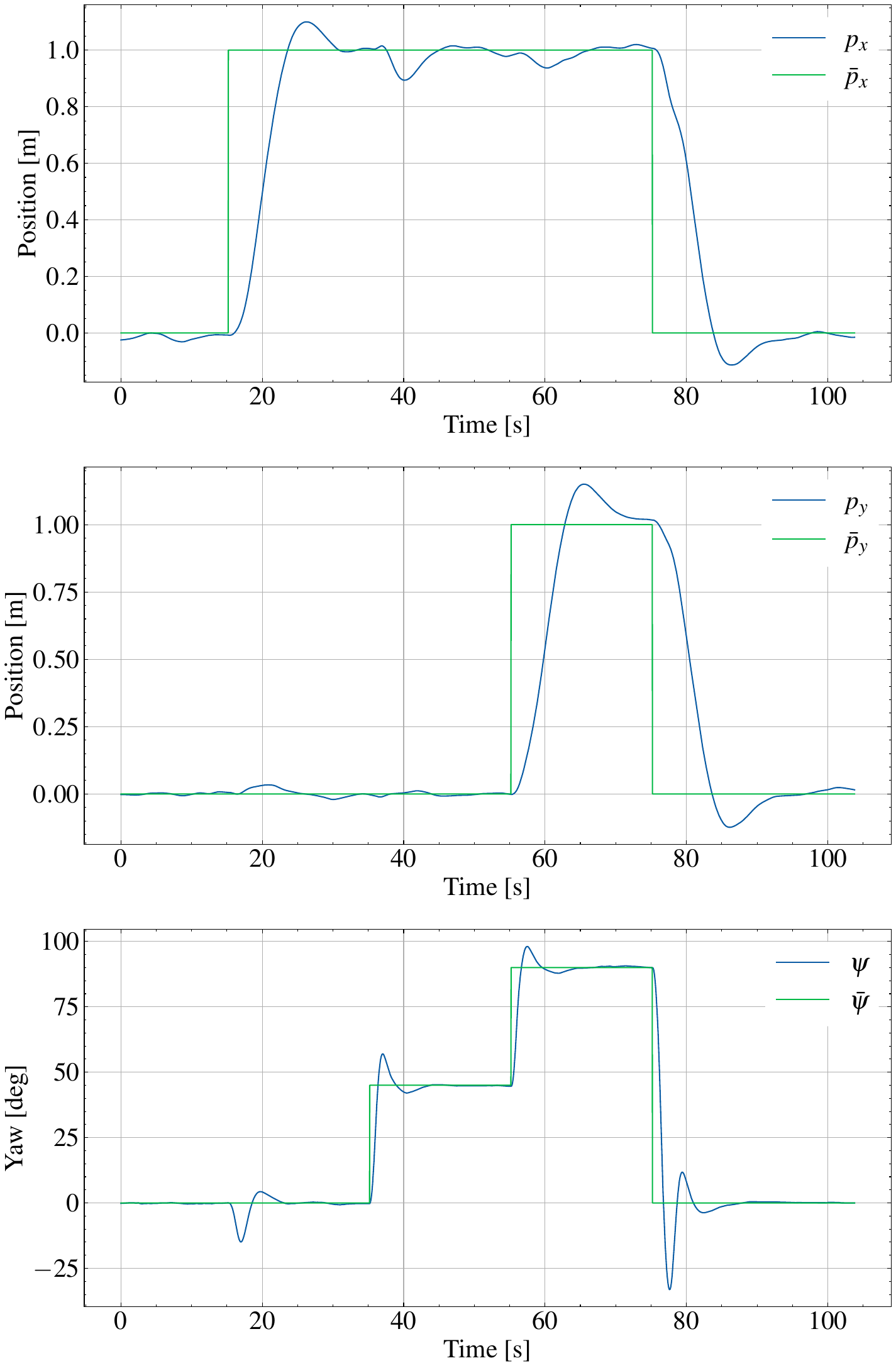}
        \caption{ATMOS \ac{SITL} performance with force and torque setpoints.} \label{subfig:sitl_wrench_states}
    \end{subfigure} 
    \hspace{2mm}
    \begin{subfigure}[t]{0.45\linewidth}
        \centering
        \includegraphics[width=\linewidth]{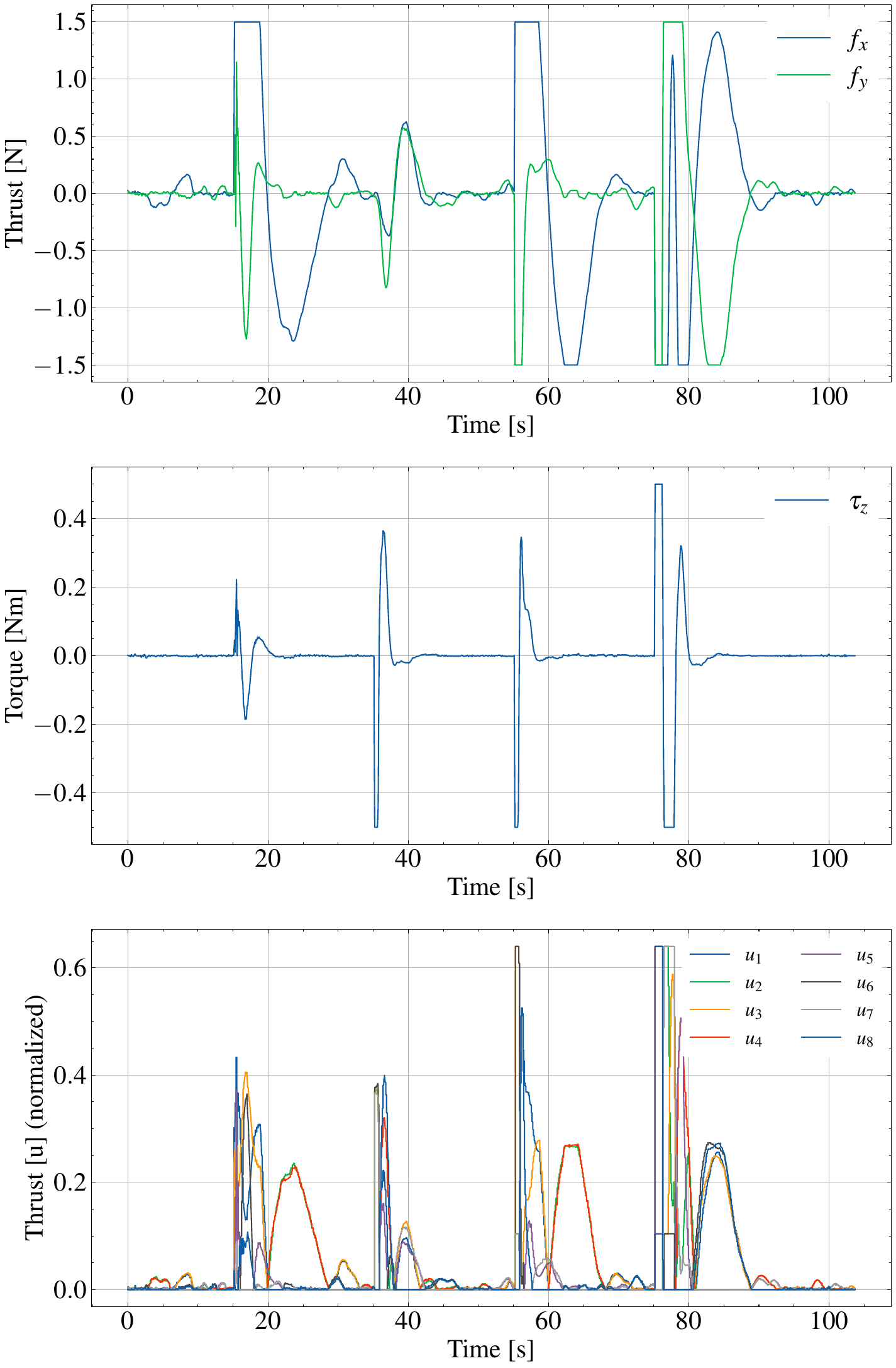}
        \caption{ATMOS \ac{SITL} force and torque control inputs.} \label{subfig:sitl_wrench_inputs}
    \end{subfigure}
    \caption{ATMOS \ac{SITL} performance when using the \ac{NMPC} scheme in \cref{ctl:wrench} and considered the Body Force and Torque model.}
    \label{fig:sitl_wrench}
\end{figure*}

In this section, we show and discuss the results achieved using some of the controllers and planners in \cref{sec:autonomy}, both in \ac{SITL} and in the real ATMOS platform. For an in-depth analysis of all results, we refer the reader to the video: \href{https://drive.google.com/file/d/1L82iIsM9dV8Sy2irFzW3wwDDPRb_ZB5r/view?usp=drive_link}{click here for the video}.

\subsection{Offboard \ac{NMPC} with ATMOS \ac{SITL}}

The \ac{SITL} simulation follows the implementation described in \cref{sec:sw_simulation}. First, we test the direct control allocation method using the model in \cref{ctl:da}. We set a sequence of multiple setpoints, spaced in time by \SI{20}{\second}, and with translations of \SI{1}{\metre} in the $x$ and $y$ axis, as well as of \SI{45}{\degree} around the $z$ axis. The results are shown in \cref{subfig:sitl_da_inputs,subfig:sitl_da_states}. In \cref{subfig:sitl_da_inputs}, we observe the normalized thrust set on each thruster $i=1,\dots,8$. We can observe in \cref{subfig:sitl_da_states} that the simulated platform can converge to the required setpoints with minimal overshoot and no steady-state error. This is expected as no external disturbances are considered in this case, and the actuation model is to the lowest level of control possible for the platform.

Then, considering the same setpoints but the body force and torque model in \cref{ctl:wrench}, we collected the results in \cref{fig:sitl_wrench}. Note that in this scenario, the \ac{NMPC} generates forces and torques that are then translated to thruster inputs. To conveniently observe this, we included in \cref{subfig:sitl_wrench_inputs} the \ac{NMPC} forces $f_x, f_y$ and torque $\tau_z$, as well as normalized inputs to each thruster (noting that a value of $1.0$ equals maximum thrust $\bar{f}$). To avoid saturation, we limited the maximum forces to $f_{max}=\SI{1.5}{\N}$ on each axis. We may also observe that for half the maximum force available per axis, the resulting input to some thrusters is considerably larger than half of the maximum thrust, particularly when translation and attitude changes are required. As the maximum force per axis is smaller than in the direct allocation case, we expect the translation to have a slower transient than in the direct allocation scenario, as can be seen when comparing \cref{subfig:sitl_da_states} to \cref{subfig:sitl_wrench_states}. Lastly, the planning section will demonstrate the results of control with rate setpoints.

\begin{figure*}[tpb]
    \centering
    \begin{subfigure}[t]{0.45\linewidth}
        \centering
        \includegraphics[width=\linewidth]{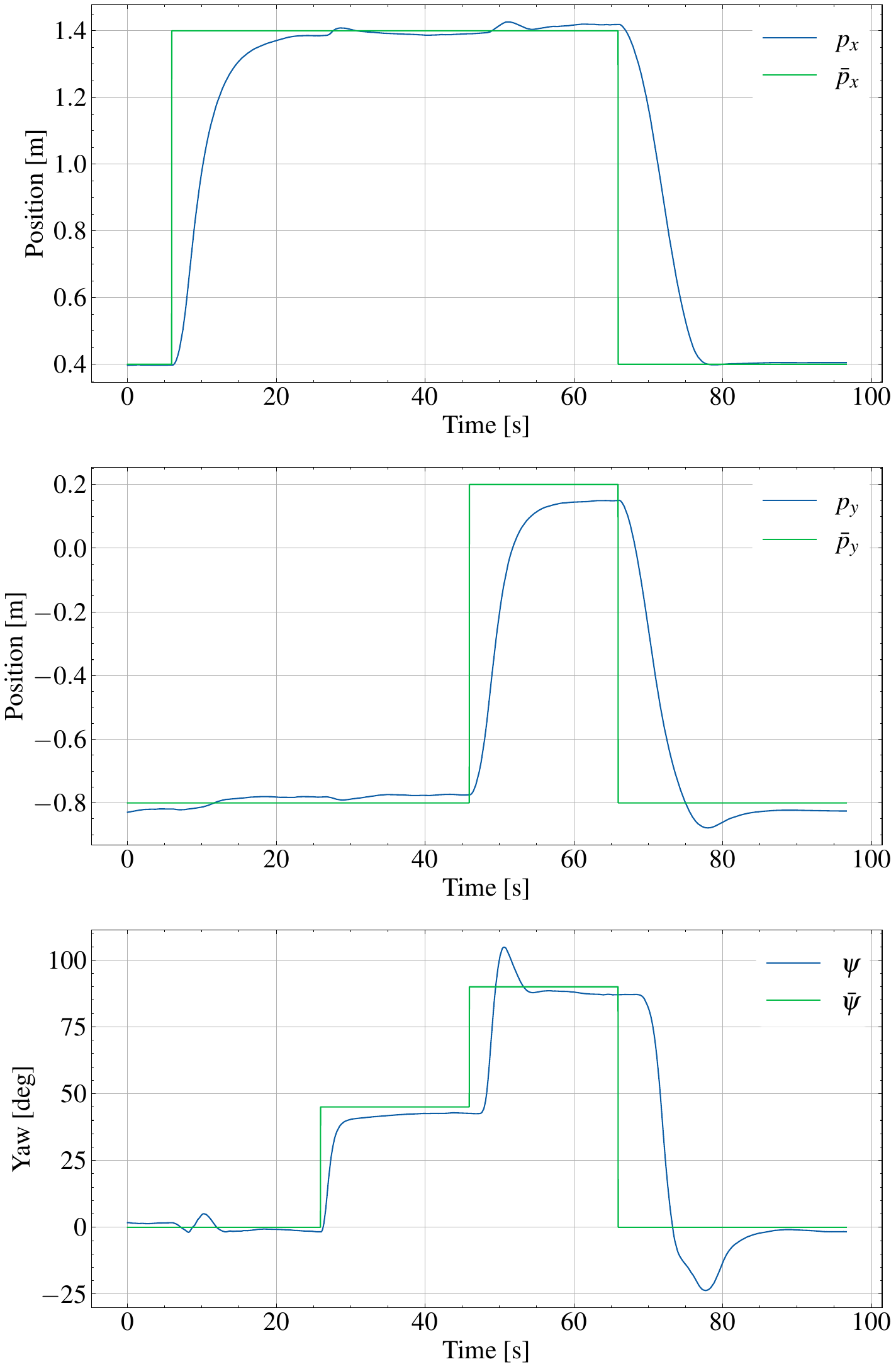}
        \caption{ATMOS hardware performance with wrench setpoints.} \label{subfig:real_wrench}
    \end{subfigure}
    \hspace{2mm}
    \begin{subfigure}[t]{0.45\linewidth}
        \centering
        \includegraphics[width=\linewidth]{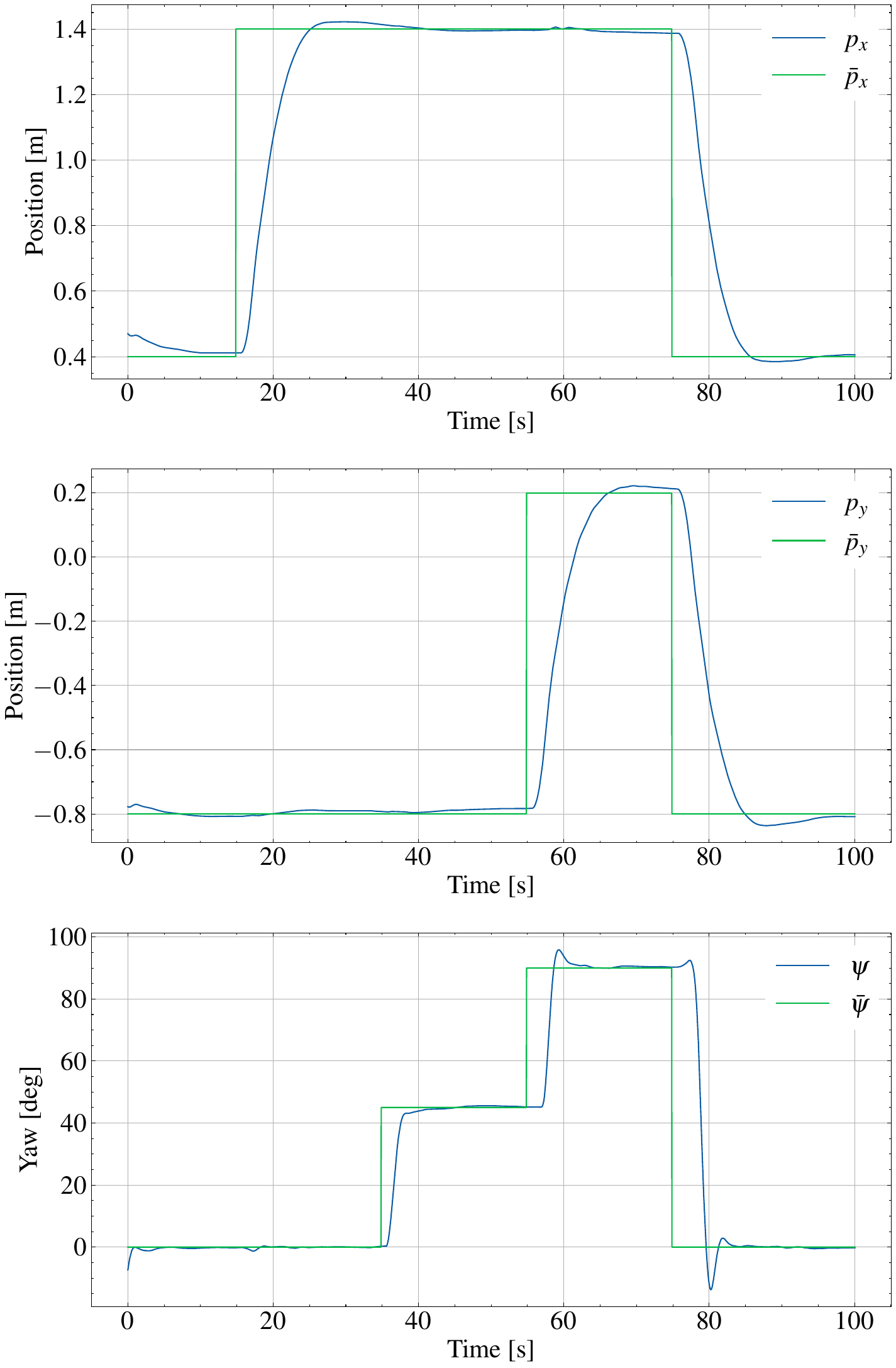}
        \caption{ATMOS hardware performance with offset-free NMPC using wrench setpoints.} \label{subfig:real_of_wrench}
    \end{subfigure}\\
    \vspace{2mm}
    \begin{subfigure}[t]{0.45\linewidth}
        \centering
        \includegraphics[width=\linewidth]{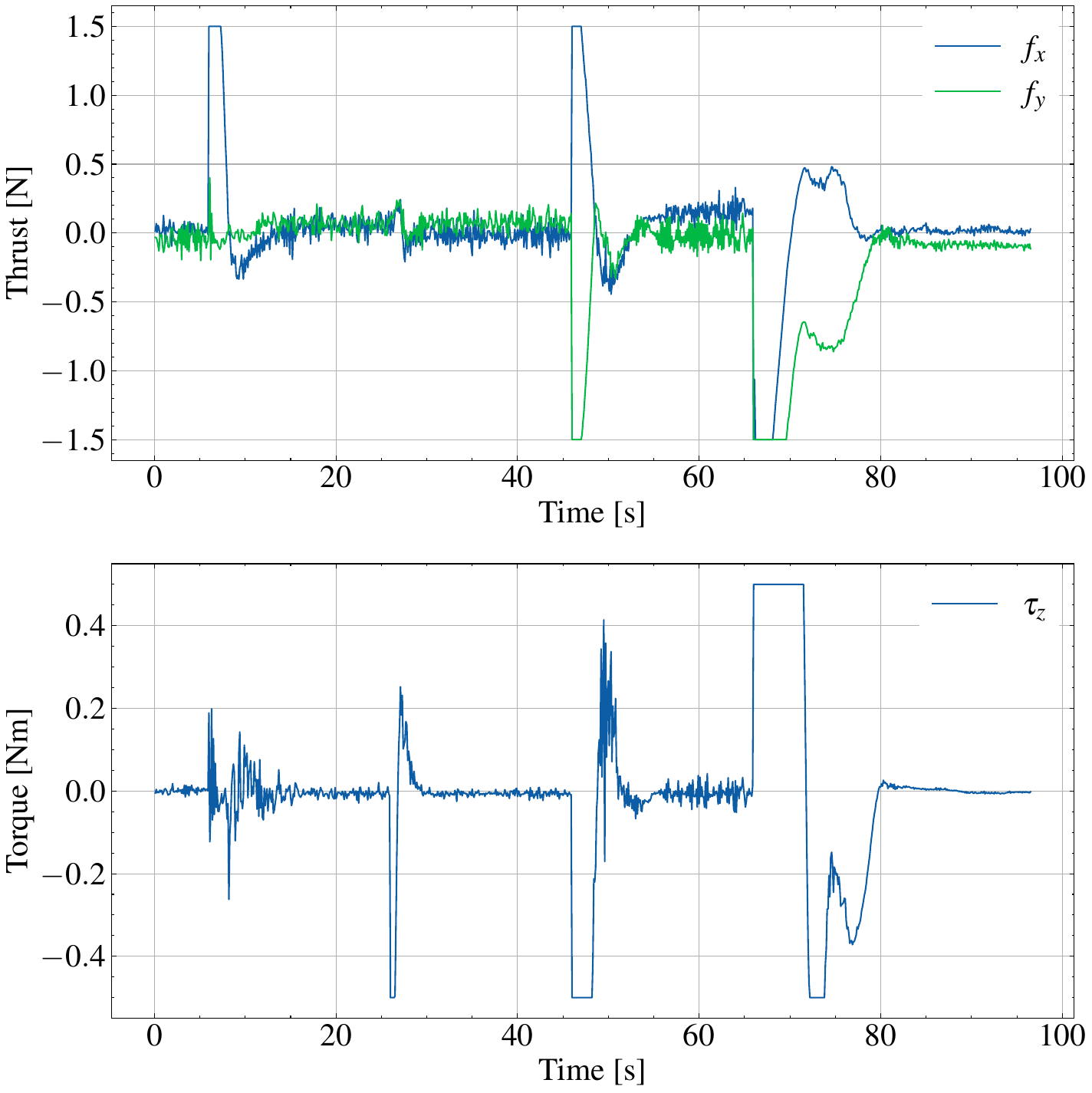}
        \caption{ATMOS hardware control inputs with wrench setpoints.} \label{subfig:real_wrench_inputs}
    \end{subfigure}
    \hspace{2mm}
    \begin{subfigure}[t]{0.45\linewidth}
        \centering
        \includegraphics[width=\linewidth]{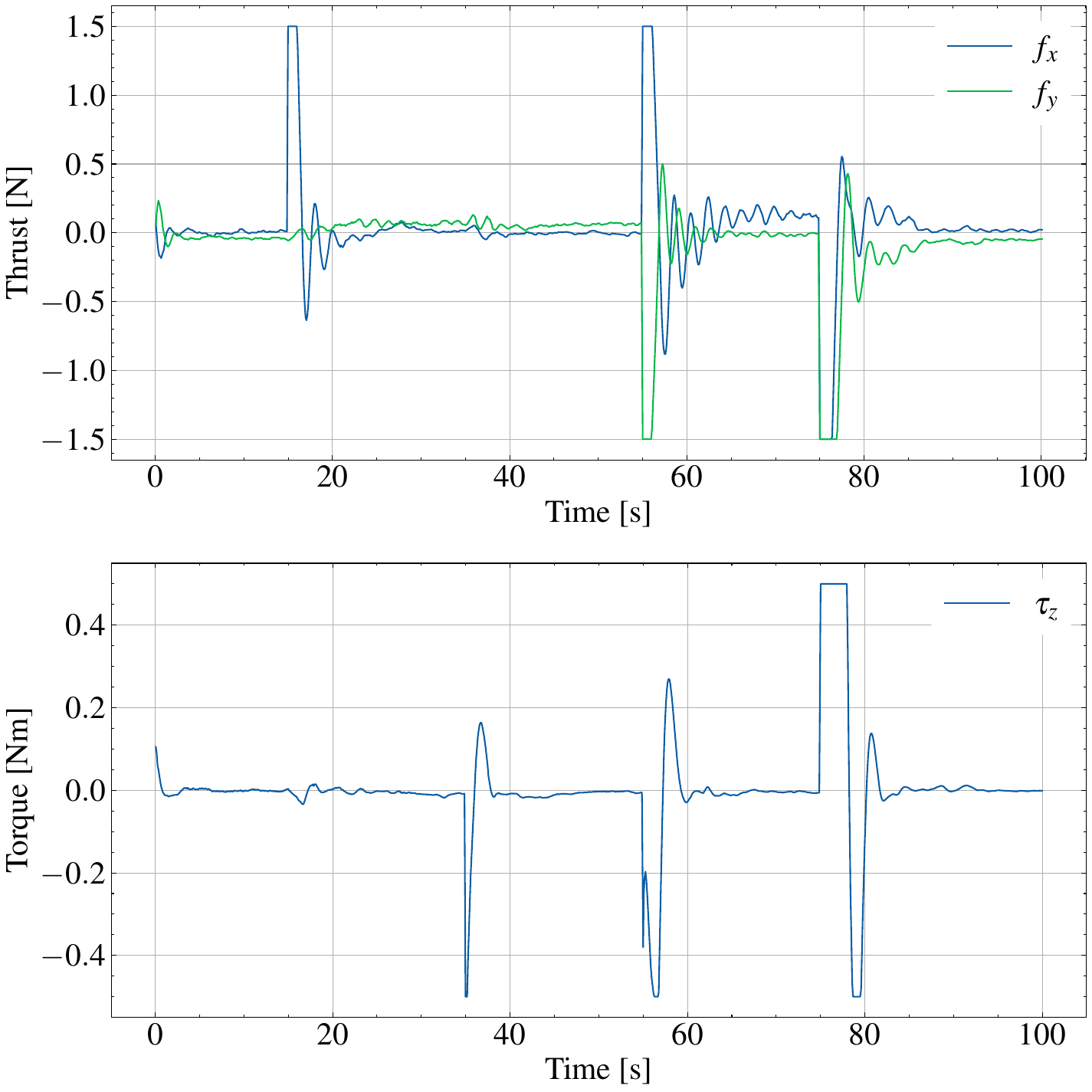}
        \caption{ATMOS hardware offset-free NMPC control inputs with wrench setpoints.} \label{subfig:real_of_wrench_inputs}
    \end{subfigure}
    \caption{ATMOS hardware performance when using the \ac{NMPC} scheme in \cref{eq:centralized_mpc} with wrench setpoints. On the left, using the nominal model in \cref{ctl:wrench}, and to the right, considering the offset-free NMPC with the model in \cref{ctl:disturbance_wrench}.}
    \label{fig:real_wrench_w_wo_offset}
\end{figure*}

\subsection{Offboard \ac{NMPC} with ATMOS Hardware}

After performing simulations on the \ac{SITL} simulator, we implemented the methods on the ATMOS platform. The experimental setup is very similar to the simulated one, with the addition of \ac{mocap} for ground-truth. 

In \cref{subfig:real_da_states} and \cref{subfig:real_da_inputs}, we present the hardware results of direct control allocation with ATMOS. In this scenario, we use the discretized version of the model in \cref{ctl:da}. The platform can track the desired references, but it is also possible to observe a steady-state error among some of the setpoints. The magnitude of the error is approximately \SI{10}{\cm} in position and \SI{5}{\degree} in attitude. We identify three sources for such error: i) floor unevenness, ii) inertial parameters mismatch, and iii) actuation model mismatch. Considering the first source, we can infer that an uneven floor will induce a constant residual force on the platform. Since such force is not in the model, it cannot be compensated in the current \ac{NMPC} framework. These forces would cause the platform the have a steady-state error with a constant input value that is reciprocal to the disturbance effect. Possible solutions to mitigate this source of error are using offset-free \ac{NMPC} schemes \cite{morari2012nonlinear}, similar to adding integral action to the controller. Regarding the second source of error, we note that during operation, both the mass of the platform and inertia change as the platform loses mass due to using the onboard propellant. Although this might change the transient behavior of the platform and cause overshooting or undershooting, it would not be sufficient to cause steady-state errors. Lastly, the actuation model considered for the \ac{NMPC} scheme is rather simplistic and does not consider losses in efficiency when triggering more than a single thruster at each sampling time. Such an event will yield a lower input than expected, causing the system to move or rotate slower than the predicted model. The effect would be similar to a wrong inertial parameter estimate. It is worth noting that despite the performance difference discussed previously, moving from a simulated environment to a real platform provided similar results, with relatively accurate transient behavior. In particular, the performance on the $x$-axis was largely unaffected (as these setpoints were placed in flatter floor areas), including during transient response. 

To overcome the effect of external disturbances, we tested the Offset-free \ac{NMPC} scheme shown in \cref{ssec:of_nmpc} using the model in \cref{ctl:disturbance_wrench}. This control scheme is compared against the nominal \ac{NMPC} with the model in \cref{ctl:wrench}. The results can be seen in \cref{fig:real_wrench_w_wo_offset}. Comparing the two columns, it is possible to observe the effect of the offset-free compensation, particularly on position $p_y$ and attitude $\psi$. Through the inclusion of the \ac{EKF} estimator, the external disturbances $d_v$ and $d_\omega$ are estimated online and the \ac{NMPC} scheme counteracts its effect, resulting in a zero steady-state error. In the generated control signals, we can observe that in \cref{subfig:real_of_wrench_inputs} there exists a large steady-state input, particularly in the \SI{60}{\second}-\SI{70}{\second} interval, to allow the system to compensate the estimated external disturbance. 

\begin{figure*}[tpb]
    \centering
    \includegraphics[width=\textwidth]{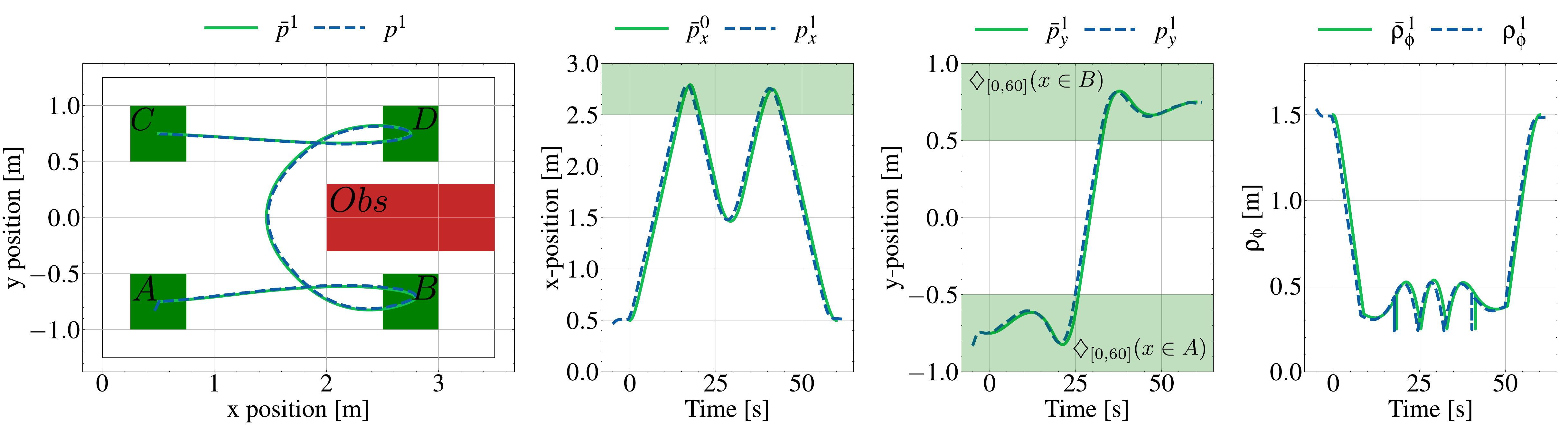}
    \caption{ATMOS hardware performance for B\'ezier trajectory tracking of a single robot with rate control.}
    \label{fig:planner_single}
\end{figure*}

\begin{figure*}[tpb]
    \centering
    \includegraphics[width=\textwidth]{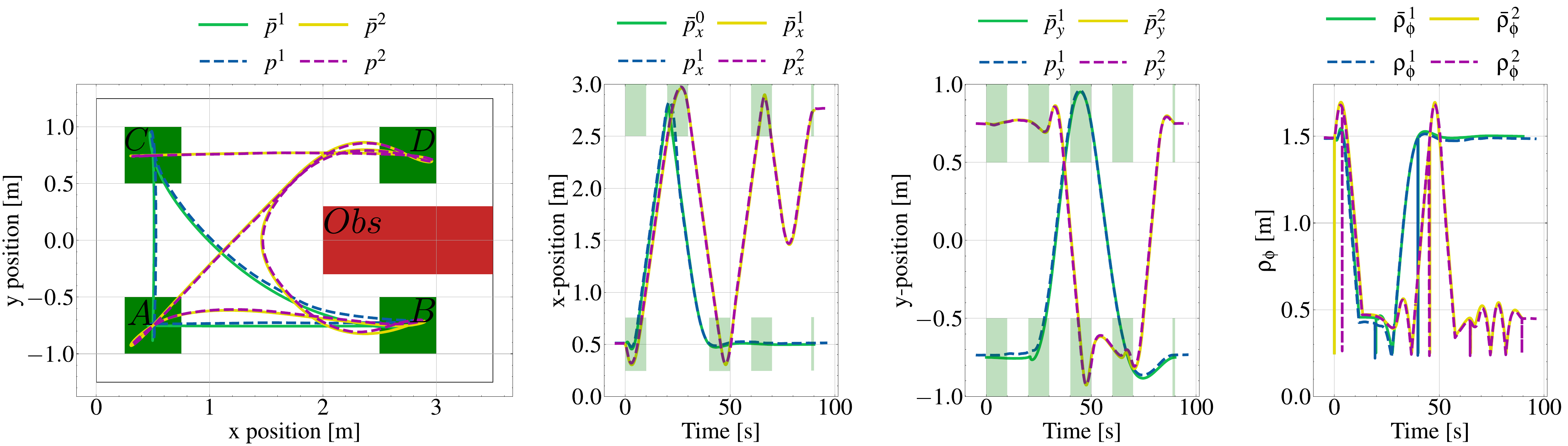}
    \caption{ATMOS hardware performance for B\'ezier trajectory tracking of two robots with rate control.}
    \label{fig:planner_multi}
\end{figure*}

\subsection{Planning}
We present the tracking results of a B\'ezier trajectory, satisfying a high-level specification using the B\'ezier trajectory parametrization and spatially robust STL planner from \cref{ssec:planning_schemes} with the rate controller in \cref{ctl:rate}.

\subsubsection{Single-Agent Scenario}
Consider a single robot $R^1$ with the initial and final state $p^1(t_0)=[0.5,-0.75] \in B$ and $p^1(t_f)=[0.5,0.75] \in C$ with $t_0=0$ and $t_f=60$.
The STL specification we consider is $\phi = \diamondsuit_{[0,60]}(p^1\in B) \land \diamondsuit_{[0,60]}(p^1\in D) \land \Box_I{[0,60]}(p^1 \notin Obs)$, specifying that the robot should visit regions $B$ and $D$ at any time in the horizon and should avoid the obstacle for all time in the time horizon, $t\in [t_0,t_f]$. 
The results are presented in \cref{fig:planner_single}, showing the $p^1_x$ and $p^1_y$ positions in the plane and over time as well as the planned and executed spatial robustness $\rho_{\phi}$. 
Note that the goal of the planner is to maximize $\rho_{\phi}$ which entails maximizing the lower bound of the presented robustness value over time. As the size of the regions of interest ($B$ and $D$) is limited (with a maximal distance to violating of \SI{0.25}{\metre}) the obtained spatial robustness of $\rho_{\phi} = \SI{0.25}{\metre}$ also ensures a clearing distance of \SI{0.25}{\metre} to the obstacle as it is part of the STL specification. 
If we would increase the size of the regions, the planner would consider an increased clearing distance whenever possible.

\subsubsection{Multi-Agent Scenario}
Consider now two robots, $R^1$ and $R^2$ with a specification over a horizon of $t_0=0$ to $t_f=90$.
For $R^1$, we consider the STL specification $\phi^1 = \diamondsuit_{[0,10]}(p^1\in A) \land \diamondsuit_{[20,30]}(p^1 \in B) \land \diamondsuit_{[40,50]}(p^1 \in C) \land \diamondsuit_{[60,70]}(p^1 \in A) \land \Box_{[0,90]}(p^1 \notin Obs)$ with $p^1(t_f) = [0.5,-0.75] \in A$ which requires the robot to visit region $A$, $B$, $C$, and again $A$ in order while always avoiding the obstacle.
For $R^2$, we consider a similar STL specification $\phi^2 = \diamondsuit_{[0,10]}(p^2 \in C) \land \diamondsuit_{[20,30]}(p^2 \in D) \land \diamondsuit_{[40,50]}(p^2 \in A) \land \diamondsuit_{[60,70]}(p^2 \in B) \land \Box_{[0,90]}(p^2 \notin Obs)$ with $p^2(t_f) = [2.75,0.75] \in D$ which requires the robot to visit region $C$, $D$, $A$, $B$, and again $D$ in order while always avoiding the obstacle.
The global STL specification is then $\phi = \phi^1 \land \phi^2$.
We additionally specify collision avoidance in the planner layer with implementation details in \cite{verhagen2024temporally}. 
The results are presented in \cref{fig:planner_multi} with a planned robustness value of $\rho_{\phi} = \SI{0.25}{\metre}$. 
Notice here that it is more apparent that the tracking of the B\'ezier curves lags behind the planned trajectories. While we penalize accelerations in the planner, we are not able to explicitly constrain them. 
The spatial robustness in the motion plan ensures that these tracking errors can be accommodated for w.r.t. the satisfaction of the STL specification.

\section{Discussion and Conclusions}
\label{sec:discussion}

An overview of existing space robotics research facilities and the different strategies used to achieve frictionless motion has been presented, as well as platform limitations. Based on these works, we created the KTH Space Robotics Laboratory with its modular free-flying ATMOS platforms and multiple support systems.

As a significant component of our contribution is the open-source availability of both the hardware and the software of this facility, it is important to qualitatively assess its potential impact on the community. To this end, we are maintaining an open repository of laboratories using ATMOS or PX4Space, available in \url{https://atmos.discower.io/others/atmos_in_the_wild/}. 
Due to the flexibility of PX4Space, adapting the software to other free-flyers requires only parameter adjustments regarding thruster placement and inertial parameters. It is also possible to adjust PX4Space to propeller-based actuation using the control and metric allocation modules developed in our software stack. An example of such a platform is available in \cite{roque2016space}. With the available documentation on PX4Space, step-by-step guide on building ATMOS, and openly available SITL simulator, we look forward to seeing the next users of our contribution.

From the results in \cref{sec:prel_results}, we may conclude that our goal of seamless transfer of experiments from simulation to hardware was achieved. Further improvements to the simulator will include adding floor unevenness disturbance using the measurements from \cref{exp:floor_meas_2} and thruster efficiency loss models, reducing the disparity of hardware experiments through higher fidelity simulation.

To complete these systems, some trade-offs had to be made. To achieve a large operational area at a reduced cost, the precision of granite tables was exchanged with the lower cost per area solution of using a self-leveling epoxy floor. These leveling issues are also present in other state-of-the-art facilities. On the other hand, the proposed offset-free MPC aids in overcoming such issues for setpoint stabilization. 
Regarding tracking performance, we expect that using estimators for dynamic residuals can improve the platform performance.
Another compromise was made in the 3\ac{dof} of ATMOS, versus other platforms capable of 5 and 6 \ac{dof}. 
With ATMOS, the priority was on providing a modular, low-cost platform based on commercial off-the-shelf parts. This leads to an easy-to-replicate platform, but more importantly, it is easily adaptable to different needs. Users may extend the platform with spherical air bearings to provide 5\ac{dof}, install gimbaled systems for testing small satellites, and integrate flight-certified hardware, among many other possible use cases.

Future work will involve finalizing the integration of the NASA Astrobee \ac{FSW} in ATMOS, integrating floor disturbance plugins and thruster efficiency loss models, and implementing collaborative load transportation strategies and fault-tolerant control schemes. We also aim to finalize the integration of our software with PX4-Autopilot to allow easier access to the software and improved long-term support.

\bibliographystyle{IEEEtran}  
\bibliography{main}

\end{document}